\documentclass[journal]{IEEEtran}
 
\newif\ifanon
\anonfalse          
 
\usepackage[T1]{fontenc}
\usepackage[utf8]{inputenc}
\usepackage{amsmath,amssymb}
\usepackage{graphicx}
\usepackage{xcolor}
\usepackage{booktabs}
\usepackage{array}
\PassOptionsToPackage{hyphens}{url}
\usepackage{url}
\usepackage[hidelinks]{hyperref}

\usepackage{tabularx}
\usepackage{longtable}
\usepackage{enumitem}
\usepackage{pdflscape}
\usepackage{placeins}
\usepackage{tikz}
\usetikzlibrary{arrows.meta,positioning,calc}
 \usepackage{subcaption}
\definecolor{fwfill}{HTML}{1F3A5F}
\definecolor{eufill}{HTML}{E8EEF5}
\definecolor{eurule}{HTML}{9DB4CC}
 
\newcolumntype{L}[1]{>{\raggedright\arraybackslash}p{#1}}
\newcommand{\hib}{\ensuremath{\uparrow}}      
\newcommand{\lib}{\ensuremath{\downarrow}}    
\newcommand{\tb}{\ensuremath{\boxdot}}        
\newcommand{\arr}{\ensuremath{\rightarrow}}   
 
\DeclareMathOperator*{\argmax}{arg\,max}
\DeclareMathOperator{\softmax}{softmax}
 
\ifanon
  
\else
  
\fi
 
\begin{document}
 
\title{A Unified Evaluation Framework for Trustworthy Large Language Models, Agentic AI, and Multimodal Systems}

\ifanon
\else
  \author{Shaina Raza,
          Ahmed Y. Radwan,
          Imran Liaquat,
          and Kathryn Hume
  \thanks{S. Raza, A. Y. Radwan, I. Liaquat, and K. Hume are with the
  Vector Institute for Artificial Intelligence, Toronto, ON M5G 0C6, Canada.
  Email: \{shaina.raza, ahmed.radwan, imran.liaquat, kathryn.hume\}@vectorinstitute.ai}%
  \thanks{Corresponding author: Shaina Raza.}}
\fi

\ifanon
\else
  \markboth{}%
  {Raza \MakeLowercase{\textit{et al.}}: A Unified Evaluation Framework for Trustworthy LLMs, Agentic AI, and Multimodal Systems}
\fi

\maketitle

\begin{abstract}
Benchmark scores alone provide an incomplete basis for assessing the trustworthiness of modern artificial intelligence systems. Large language models (LLMs), agentic systems, and multimodal models (MLLMs) require different forms of assessment, yet their evaluation evidence must remain interpretable for development and oversight. We propose a unified framework that connects output-level, trajectory-level, and cross-modal assessment through eight trustworthiness dimensions: capability, robustness, safety, fairness, transparency, governance, oversight, and efficiency. The framework preserves system-specific metrics while mapping native measurements to common performance bands, accompanied by uncertainty estimates and traceable evidence. A meta-evaluation layer examines the validity, reliability, and reproducibility of the evaluation itself. Multidimensional profiles expose strengths and weaknesses, while safety-critical overrides prevent aggregate scores from masking critical failures. Mappings to governance frameworks, international standards, and European Union regulatory requirements connect technical assessment with oversight needs. The framework provides a structured basis for assessing both system performance and the credibility of the evidence supporting it, with empirical validation across deployment contexts remaining an essential next step.
\end{abstract}

\begin{IEEEkeywords}
AI evaluation, large language models, agentic systems, multimodal models,
trustworthy AI, benchmarking, AI governance, EU AI Act.
\end{IEEEkeywords}

\section{Introduction}
\label{sec:introduction}
\IEEEPARstart
{M}{odern} artificial intelligence (AI) evaluation is fragmented across model classes. Large language models (LLMs) are commonly assessed through output-level benchmarks, agentic systems require trajectory-level evaluation of planning, tool use, recovery, and oversight, and multimodal systems introduce cross-modal grounding and consistency requirements. These differences make it difficult to compare evaluation evidence across systems and to determine whether reported scores are sufficiently valid, reproducible, and informative for governance decisions. 

This article addresses this gap through a unified evaluation framework for LLM, agentic, multi-agent, and multimodal systems. The framework combines a common trustworthiness spine with model-specific assessment tracks and a meta-evaluation layer that evaluates the quality of the evaluation process itself. It is intended to support both AI developers and evaluators while providing evidence that is interpretable by governance and assurance stakeholders and the broader scientific community.

The framework covers three primary model classes: (1) LLMs, including fine-tuned open and API-served models; (2) agentic and multi-agentic systems that plan and act over multiple steps, including tool-using and retrieval-augmented (RAG) agents; and (3) multimodal models such as vision language models (VLMs), omni modality,  speech models, and models that process or generate multiple modalities. 
The framework operates primarily at the level of the deployable AI system, including its prompts, tools, interfaces, and guardrails. It can be applied during development as well as to systems already in use. Training-time decisions are outside its primary scope except where they affect observable evaluation evidence. The framework is not intended to replace legal conformity assessment, and domains for which sufficiently mature evaluation instruments are not yet defined, such as embodied robotics and real-time control, remain outside the current scope.

This framework is based on trustworthiness in AI, where trustworthiness is understood as the ability to meet stakeholders' expectations in a verifiable manner. Consistent with ISO/IEC TS 5723:2022 \cite{iso5723}, the characteristics relevant to trustworthiness depend on the system and service being evaluated, the technologies, data, and processes involved, and the applicable context or sector. The overarching goal of this work is therefore to operationalize this context-sensitive concept through eight evaluation dimensions and the supporting evidence required to make evaluation results independently interpretable and verifiable.

\textbf{Contributions}
This article makes three main contributions. (1) It introduces a unified eight-dimensional evaluation spine for LLMs, agentic, multi-agent, and multimodal systems while preserving their distinct units of assessment: outputs, trajectories, and cross-modal relationships. (2) It develops a common reporting architecture that combines native metrics with normalization, uncertainty reporting, not-applicable rules, safety-critical overrides, and a meta-evaluation layer for assessing the validity and reproducibility of the evaluation itself. (3) It operationalizes the framework through model-specific evaluation instruments and evidence requirements and maps these components to AI governance and regulations, such as EU AI Act, ISO standards and various AI Risk Management Framework (RMF) \cite{nistairmf}.

\begin{figure*}[h]
    \centering
    \includegraphics[width=0.8\linewidth]{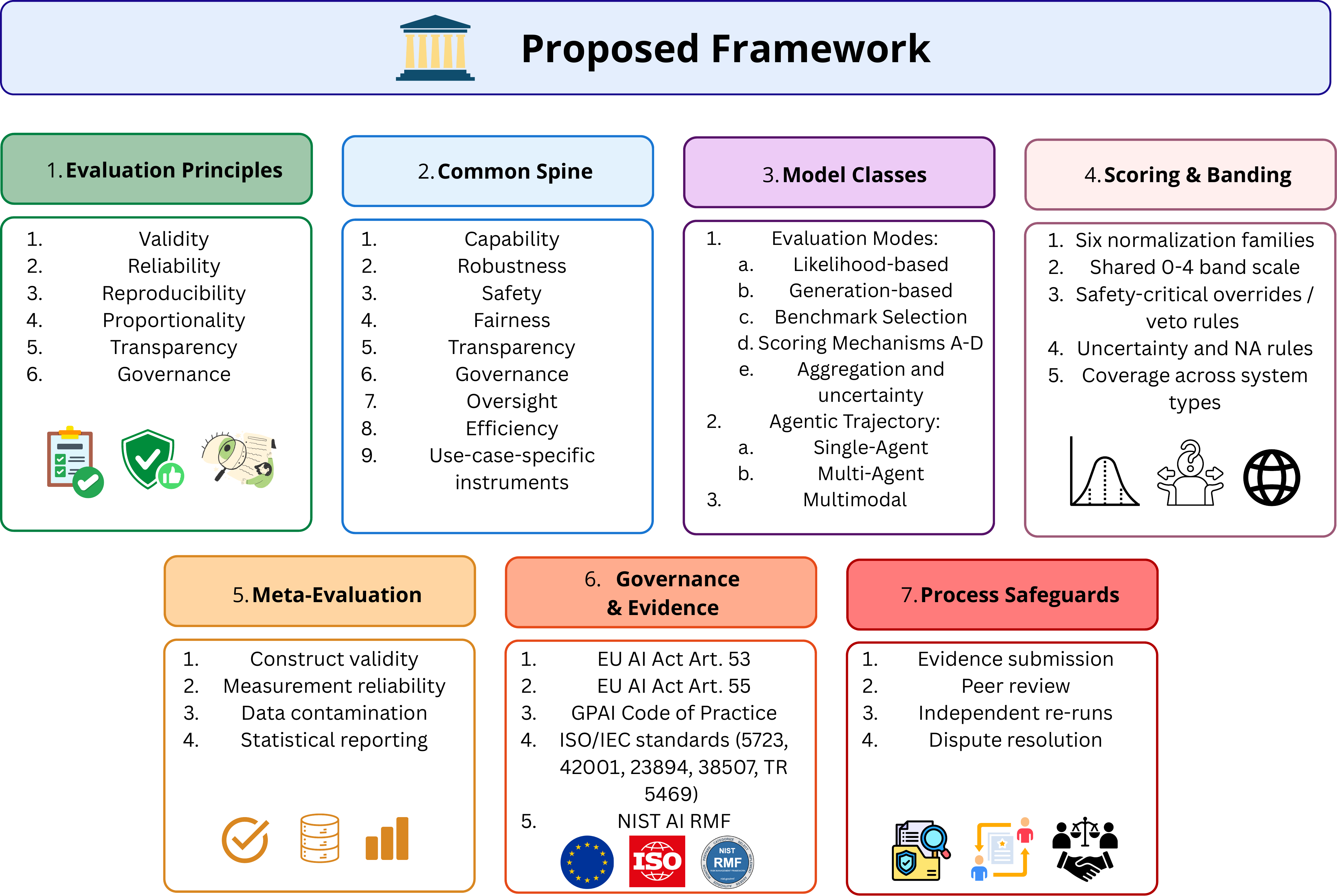}
\caption{Common evaluation dimensions in the framework.}
    \label{fig:aixpert-framework}
\end{figure*}

\section{Related Work}

\textbf{Evaluation Frameworks for Language Models}
Evaluation of LLMs has traditionally focused on individual capabilities or failure modes through task-specific benchmarks. Common evaluation dimensions include factuality, reasoning, knowledge, code generation, toxicity, adversarial robustness, and calibration. Benchmarks such as TruthfulQA \cite{lin2022truthfulqa} and HaluEval \cite{li2023halueval} target factuality and hallucination; GSM8K \cite{cobbe2021gsm8k}, BIG-Bench Hard \cite{suzgun2023bbh}, and ARC-Challenge \cite{clark2018arc} assess reasoning; MMLU and MMLU-Pro \cite{wang2024mmlupro} measure broad knowledge. Safety-oriented evaluation is commonly conducted using datasets such as   ToxiGen \cite{hartvigsen2022toxigen}, HarmBench \cite{mazeika2024harmbench}, and others. 

Multimodal evaluation needs to assess both individual modalities and their interactions. Existing benchmarks cover visual question answering (VQA), captioning, retrieval, grounding, OCR, speech recognition, and cross-modal hallucination using datasets such as VQAv2 \cite{goyal2017vqav2}, MS-COCO \cite{lin2014coco}, HumaniBench \cite{raza2025humanibench} and others.  These benchmarks are largely task-specific and use different native metrics, comparing trustworthiness properties across multimodal, text-only, and agentic systems remains difficult. 

\textbf{Agentic and Multi-Agent Evaluation}
 Agentic AI systems introduce evaluation requirements that extend beyond the correctness of a final response. Their behaviour unfolds through trajectories involving planning, tool use, intermediate decisions, environment interactions, and recovery from errors. Consequently, recent agent evaluation work \cite{farooq2026evaluating} increasingly considers task completion, tool-use correctness, trajectory efficiency, error recovery, and adherence to constraints in addition to final task success.
These efforts demonstrate that agentic evaluation must move from output-level assessment toward trajectory- and interaction-level analysis. However, agentic metrics are often developed independently of conventional LLM evaluation and broader governance requirements. 

\textbf{Regulatory and Standards-Based Assessment}
A parallel body of work addresses AI evaluation from a governance and risk-management perspective. The European Union Artificial Intelligence Act 2026 \cite{eu_ai_act_2026} establishes obligations related to documentation, transparency, model evaluation, adversarial testing, risk management, cybersecurity, and incident reporting, with additional requirements for general-purpose AI models presenting systemic risk. The GPAI provides further guidance for implementing these obligations.
International standards provide complementary perspectives. ISO/IEC TS 5723 \cite{iso5723} defines AI trustworthiness concepts; ISO/IEC 42001 \cite{iso42001} addresses organizational AI management systems; ISO/IEC 23894 \cite{iso23894} provides guidance for AI risk management; and ISO/IEC 38507 \cite{iso38507} addresses governance of organizational AI use. The NIST AI RMF similarly organizes AI risk management around the GOVERN, MAP, MEASURE, and MANAGE functions.
These frameworks provide important governance structures, but they typically do not prescribe a unified set of operational metrics spanning language, agentic, multi-agent, and multimodal systems. Conversely, benchmark-oriented evaluation research often provides detailed metrics without connecting them systematically to governance evidence and assurance processes.

\textbf{Positioning of This Work}
Rather than introducing another benchmark suite, we organize existing and emerging instruments around eight trustworthiness dimensions. LLMs are assessed at the output level, agentic and multi-agent systems at the trajectory level, and multimodal systems at the cross-modal level. A meta-evaluation layer further examines validity, reliability, contamination, and statistical quality, enabling heterogeneous measurements to remain model-appropriate while supporting a common evaluation and governance framework.
 
\section{Evaluation Principles}
\label{sec:principles}

Our proposed framework rests on six principles. They define what constitutes a rigorous measurement, and every scoring decision in Section~\ref{sec:framework} is guided by these principles. 
\begin{enumerate}
  \item \textbf{Validity:} each metric measures the construct it is intended to assess.
  \item \textbf{Reliability:} measurements are stable across runs and raters. This principle concerns measurement reliability; reliability as a property of the system under evaluation is assessed under Robustness (Table~\ref{tab:common_dimensions}).
  \item \textbf{Reproducibility:} sufficient provenance is recorded to support independent reproduction.
  \item \textbf{Proportionality:} the depth of evaluation scales with the model capability and risk, consistent with the EU's risk-based approach (Section~\ref{sec:eu}).
  \item \textbf{Transparency:} assumptions, limitations, and negative results are reported.
  \item \textbf{Governance:} every reported result is traceable to an accountable source and reproducible from an auditable evidence trail, and each model is accompanied by its governance artifacts (documentation, training-data summary, access and incident controls).
\end{enumerate}

\subsection{From Native Scores to a Shared Band}
In this framework, the scoring proceeds in two steps. Each benchmark is first evaluated with its own native method, and the native score is then mapped to a common 0 - 4 band to support consistent interpretation across heterogeneous instruments. The mapping rule depends on the metric and is applied by one of six normalization families. The families, their thresholds, and the per-metric assignments are defined in Section~\ref{sec:banding} and specified in each descriptor  (Appendix~\ref{app:catalog}), so they are not restated here.

\begin{center}
\begin{tabular}{cl}
\toprule
\textbf{Band} & \textbf{Reading} \\
\midrule
0 & Unacceptable \\
1 & Below expectations \\
2 & Adequate \\
3 & Good \\
4 & Excellent \\
\bottomrule
\end{tabular}
\end{center}

\subsection{The Eight-Dimensional Profile}

\begin{figure}[h]
    \centering
    \includegraphics[width=\linewidth]{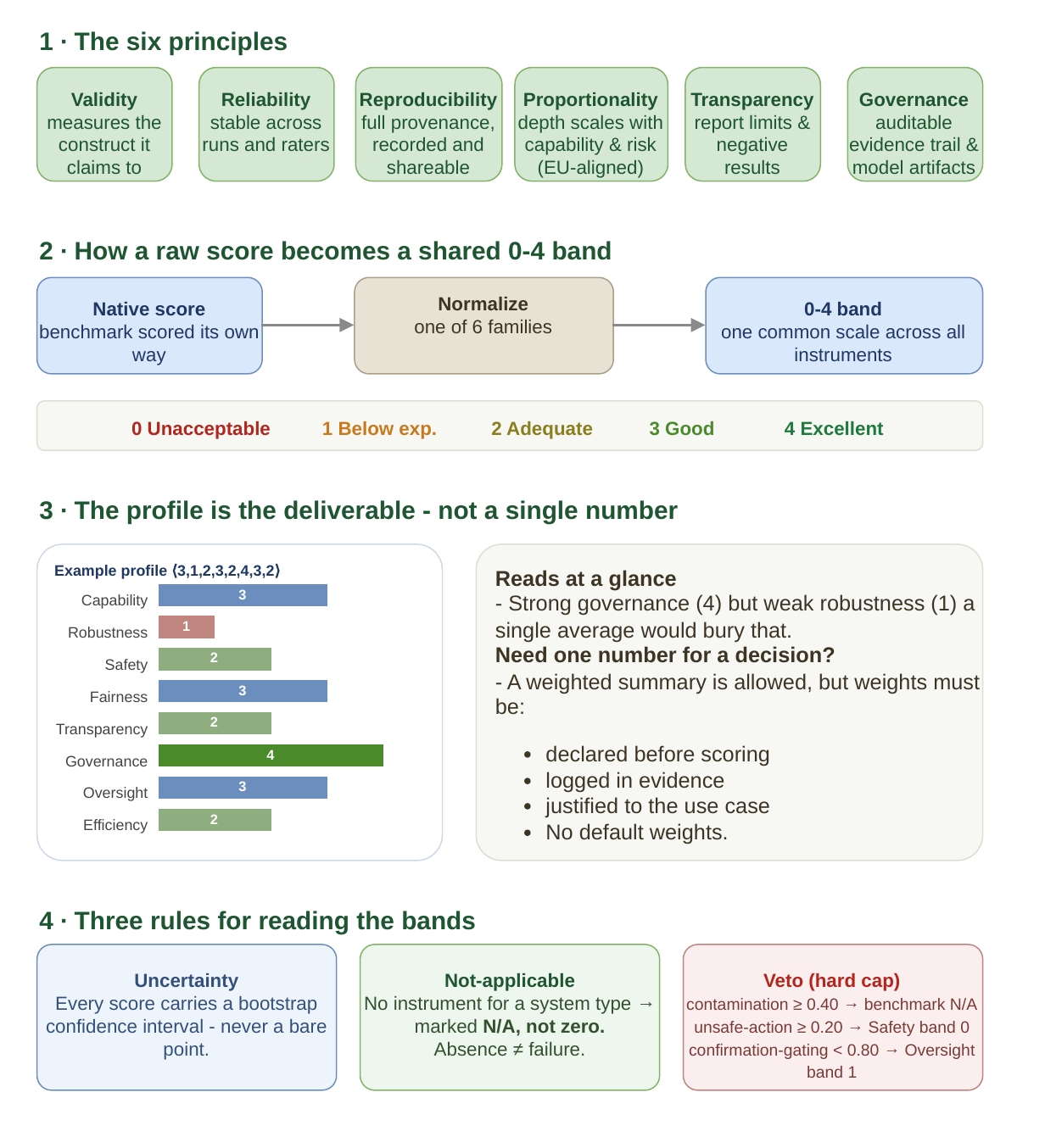}
    \caption{\textbf{The evaluation profile}. \textbf{Panel~1}: the six measurement principles. \textbf{Panel~2}: the two-step path from a native benchmark score to the shared 0 - 4 band. \textbf{Panel~3}: the eight-dimensional profile, reported per dimension rather than collapsed into a single average. \textbf{Panel~4}: the three rules governing reported bands (uncertainty, not-applicable, and veto).}
    \label{fig:profile}
\end{figure}

We report an eight-dimensional profile (Figure~\ref{fig:profile}) rather than aggregating all dimensions into a single score. A single aggregate score may be disproportionately influenced by capability benchmarks and could hide trade-offs such as strong performance paired with weaker safety. A stakeholder interpreting the results, whether a downstream integrator, an EU reviewer, or a consortium decision board, interprets the profile at the level of individual dimensions.
When a specific decision needs a summary, for example, a deployment decision, a weighted aggregation may be computed, but the weights must be declared before scoring, recorded in the evidence package, and justified with respect to the intended use case. The framework prescribes no default weight vector, because the appropriate weights depend on context.

\subsubsection{Interpretation of Band Scores}

Three rules govern the reported bands, and each is defined in full in Section~\ref{sec:banding}.

\begin{description}
  \item[Uncertainty] Every score carries a bootstrap confidence interval, rather than a point estimate alone.
  \item[Not-applicable] A dimension with no applicable instrument for a given system type is marked \textit{N/A} rather than scored zero, so absence of measurement is not interpreted as failure (coverage matrix, Table~\ref{tab:coverage}).
\item[Veto] Three safety-critical instruments override normal banding. Contamination $\geq 0.40$ voids the affected benchmark, whose construct is then marked \textit{N/A} for that submission. Unsafe-action rate $\geq 0.20$ caps the Safety dimension at band~0, and confirmation-gating $< 0.80$ caps the Oversight dimension at band~1 (Section~\ref{sec:banding}).
\end{description}

Model outputs can be evaluated using two scoring modes: from its log-probabilities (likelihood-based, deterministic) or from its generated text (generation-based, decoding-dependent). The default protocol prioritizes likelihood-based scoring when available. The rule, its rationale, and the mode each instrument uses are specified in Section~\ref{sec:llm}.

 \section{Unified Evaluation Framework}
\label{sec:framework}
The proposed framework evaluates LLM, agentic, multi-agent, and multimodal systems using a common eight-dimensional profile, model-specific evaluation tracks, and a meta-evaluation layer that assesses the quality of the evaluation process itself. The goal is to support consistent interpretation and independent scrutiny of reported results. 
The following sections describe each component in detail.

 \subsection{Common Evaluation Spine}
\label{sec:spine}
The framework adopts a \emph{common spine} of evaluation dimensions applied uniformly to every model, as listed in Table~\ref{tab:common_dimensions}. The spine is fixed rather than selected on a per-model basis, ensuring that systems with different modalities, capabilities, and deployment contexts remain interpretable within a common evaluation structure and that each evaluation dimension is explicitly considered.
The spine defines \emph{what} must be evaluated, not \emph{how} it should be measured. 
Each dimension is operationalized using evaluation instruments appropriate for the model under assessment, allowing the underlying metrics to vary while preserving a consistent evaluation structure. Safety and Governance follow a principle of \emph{proportionality}: the rigor and extent of evidence required increase with both model capability and deployment risk, but every dimension is considered for every system, with non-applicable dimensions explicitly marked N/A. 

Results are reported as a multidimensional profile rather than a single aggregate score, preventing strong performance in one area from obscuring weaknesses in others. Across the framework, accuracy is treated as a general measurement concept rather than as a single evaluation instrument. It is defined as the measure of closeness of results of observations, computations, or estimates to the true values or the values accepted as being true~\cite{iso17572part1}. Each accuracy-based metric operationalizes this definition against the reference labels, reference outputs, task-specific oracle, or environment state.

\paragraph{Use-Case-Specific Evaluation}
The common spine remains the same across systems, but some evaluation needs depend on how and where an AI system is used. This is especially important for agentic AI systems operating over longer, multi-turn interactions. For example, a personal-care agent may need to be evaluated on whether it escalates control to a human operator when a user is in distress, while an educational agent may need to resist sycophancy and maintain evidence-based responses throughout a conversation. The framework therefore allows additional use-case-specific instruments to be added under the existing eight dimensions, without changing the common spine.
The evaluation coverage for each model category is summarized in the coverage matrix (Table~\ref{tab:coverage}), while detailed definitions and metric specifications for every dimension are provided in the catalogue in Appendix~\ref{app:catalog}.

\begin{table*}[t]
\footnotesize
\centering
\caption{Common evaluation dimensions in the proposed framework.}

\label{tab:common_dimensions}
\begin{tabularx}{\textwidth}{@{}p{0.12\textwidth} X@{}}
\toprule
\textbf{Dimension} & \textbf{What it captures} \\
\midrule
Capability & Core task performance on representative workloads. The dimension operationalizes capability, defined as a measure of capacity and the ability of an entity to achieve its objectives~\cite{iso5723}. \\
\cmidrule(lr){2-2}
Robustness & Consistency under repetition, distribution shift, and non-adversarial perturbation. The dimension operationalizes robustness, defined as the ability of a system to maintain its level of performance under a variety of circumstances~\cite{iso22989}, and covers reliability, the property of consistent intended behaviour and results~\cite{iso27000}. Adversarial pressure is scored under Safety. \\
\cmidrule(lr){2-2}
Safety & Harmful outputs or actions, misuse potential, adversarial and jailbreak resistance, dangerous-capability risks, and privacy risks such as data leakage or unauthorized disclosure. The dimension covers privacy, freedom from intrusion into the private life or affairs of an individual~\cite{iso2382}, and security, resistance to intentional, unauthorized act(s) designed to cause harm or damage to a system~\cite{iso23643}. Evaluation depth is scaled by proportionality. \\
\cmidrule(lr){2-2}
Fairness & Disparate performance or harm across groups and contexts. \\
\cmidrule(lr){2-2}
Transparency & Clarity of model documentation, trustworthiness of reported confidence, and the ability to provide understandable and evidence-grounded explanations of model outputs, decisions, or actions. The dimension operationalizes transparency, defined as the open, comprehensive, accessible, clear and understandable presentation of information~\cite{iso5723}. \\ \\
\cmidrule(lr){2-2}
Governance & Evidence that the system, its provenance, and its operation can be independently checked, together with the documentation and controls needed for accountability. For systems, accountability is defined as the property that ensures the actions of an entity can be traced uniquely to that entity~\cite{iso7498part2}. Relevant documentation and controls include model and data documentation, version and change records, access controls, and incident-reporting processes. \\
\cmidrule(lr){2-2}
Oversight & Ability of a human to understand, intervene in, and halt the system. This includes controllability, defined as the property of a system that allows a human or another external agent to intervene in the system's functioning~\cite{iso22989}. \\
\cmidrule(lr){2-2}
Efficiency & Compute, latency, energy consumption, and other resource costs. \\
\bottomrule
\end{tabularx}
\end{table*}

\subsection{Model-Specific Evaluation Tracks}
\label{sec:model-classes}
The framework defines three model-specific evaluation tracks according to the unit of evaluation: LLMs are evaluated at the \emph{output} level, agentic systems over complete action \emph{trajectories}, and multimodal systems at the level of \emph{cross-modal relationships}. These tracks are not mutually exclusive and can therefore be combined. For example, a multimodal agent is evaluated using trajectory-level instruments for its actions and cross-modal instruments for its multimodal inputs and outputs, rather than being treated as a separate fourth class. The common spine applies in all cases, with each class adding the instruments appropriate to its evaluation unit.

\subsubsection{LLM Evaluation: Output Level}
\label{sec:llm}

\begin{figure}[t]
    \centering
    \includegraphics[width=\columnwidth]{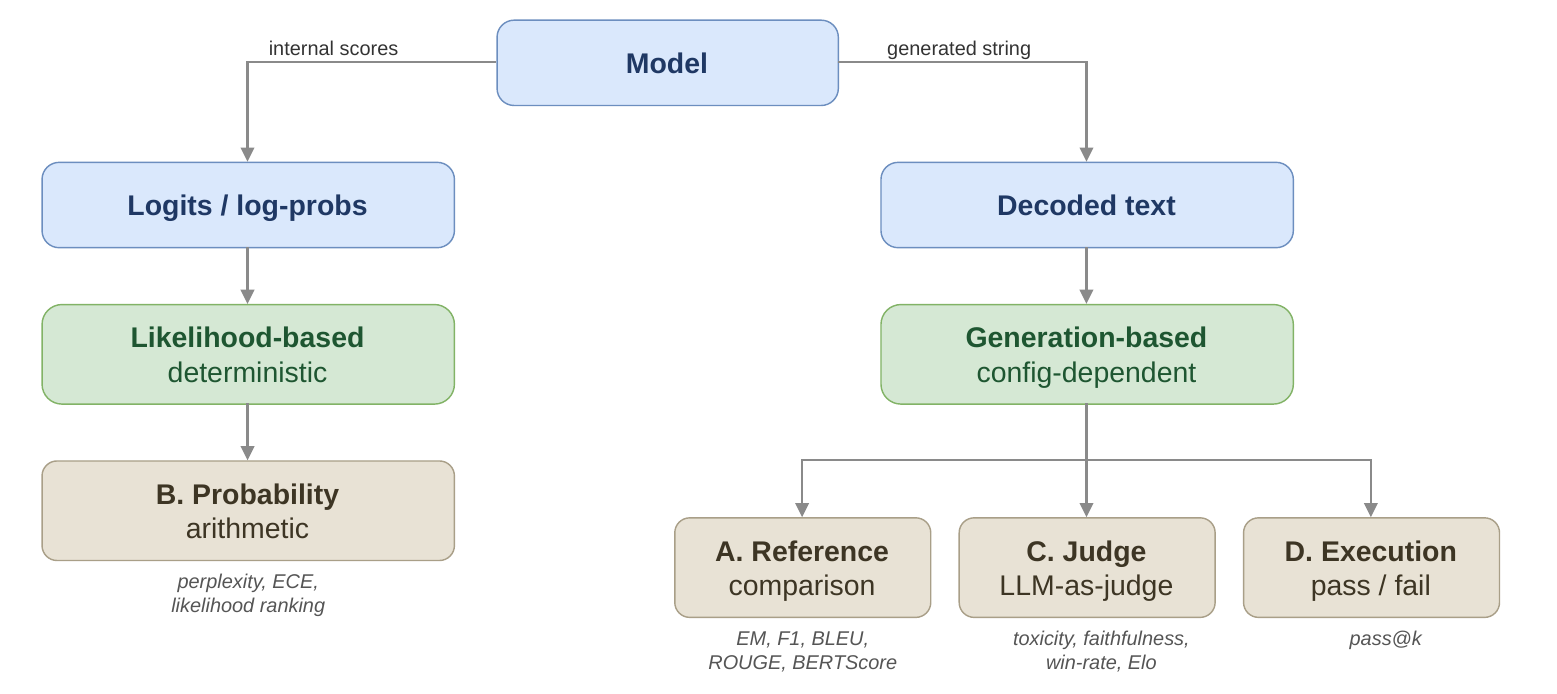}
   \caption{Output-level evaluation of LLMs. A model produces two observable
signals: internal likelihood scores (logits and log-probabilities) and
generated text. These support two evaluation modes: likelihood-based, which
is deterministic and operates directly on model scores, and generation-based,
which depends on decoding configuration. Four scoring mechanisms follow:
(A) reference comparison, (B) probability arithmetic, (C) LLM-as-judge
scoring, and (D) execution-based evaluation. Efficiency metrics and
rule-based rates are read from external instrumentation and are not shown.}
    \label{fig:llm-signals}
\end{figure}
The unit of evaluation is the model \emph{output}. A model exposes two signals, and the one that an instrument reads determines how the output is scored (Figure~\ref{fig:llm-signals}).

\paragraph{Evaluation Modes}
\begin{description}[leftmargin=0pt,itemsep=4pt]
  \item[Likelihood-based (log-probabilities).] Likelihood-based scoring operates on the model's internal probability estimates. A \emph{logit} is the raw, unnormalized score assigned to each candidate next token; the softmax turns the full set of logits into a probability distribution over the vocabulary, and the logarithm of those probabilities gives the \emph{log-probabilities}. For closed-form items (multiple choice), the prompt is paired with each candidate answer and the option with the highest log-probability is predicted; no text is generated, so scoring is deterministic and computationally efficient. This mode supports perplexity, calibration-error estimation, and likelihood-based answer ranking, and is used primarily for knowledge and multiple-choice reasoning tasks.
  \item[Generation-based (decoded text).] The string the model produces after a decoding step (greedy or sampled). The text is parsed and compared to a reference or scored by a judge: correctness against a reference answer, toxicity classifiers, an LLM-as-judge rating. Required for open-ended tasks such as chain-of-thought math, code, and summarization. Results depend on the decoding configuration (temperature, top-$p$, seed) and on an answer-extraction step.
\end{description}

\noindent The same model can score differently under the two modes, so the protocol specifies which mode each instrument uses and records the decoding configuration in the evidence package.

\paragraph{Benchmark Selection}

The output-level instruments are organized by the framework's evaluation dimensions. Within each dimension, benchmarks are selected to assess specific facets of that dimension, such as factuality, reasoning, calibration, or harmful behaviour. Table~\ref{tab:llm} provides the full metric descriptors. Where applicable, the framework identifies a primary benchmark for consistent comparison across evaluations and a secondary benchmark for additional coverage. Public benchmarks with known contamination risks are paired with contamination checks in the meta-evaluation layer.

\begin{table}[t]
\centering
\footnotesize
\caption{Representative benchmarks for LLM evaluation. MC = multiple choice.}
\label{tab:benchmarks}

\resizebox{\columnwidth}{!}{%
\begin{tabular}{llll}
\toprule
Construct & Primary & Secondary & Notes \\
\midrule
Factuality & TruthfulQA \cite{lin2022truthfulqa} & HaluEval \cite{li2023halueval} & MC + generative scoring \\
\cmidrule(lr){2-4}
Reasoning (math) & GSM8K \cite{cobbe2021gsm8k} & MATH \cite{hendrycks2021math} & Chain-of-thought scoring \\
\cmidrule(lr){2-4}
Reasoning (general) & BIG-Bench Hard \cite{suzgun2023bbh} & ARC-Challenge \cite{clark2018arc} & MC likelihood scoring \\
\cmidrule(lr){2-4}
Knowledge & MMLU \cite{hendrycks2021mmlu} & MMLU-Pro \cite{wang2024mmlupro} & Use MMLU-Pro if contaminated \\
\cmidrule(lr){2-4}
Adversarial & HarmBench \cite{mazeika2024harmbench} & AdvBench \cite{zou2023universal} & Attack success rate \\
\cmidrule(lr){2-4}
Toxicity & RealToxicityPrompts \cite{gehman2020rtp} & ToxiGen \cite{hartvigsen2022toxigen} & Classifier scored \\
\cmidrule(lr){2-4}
Code & HumanEval \cite{chen2021evaluating} & MBPP \cite{austin2021mbpp} & Pass@$k$ \\
\cmidrule(lr){2-4}
Calibration & ECE (MMLU) \cite{guo2017calibration} & -- & Requires log-probabilities \\
\bottomrule
\end{tabular}%
}

\end{table}
\noindent Candidate benchmarks and datasets, exact versions, split hashes, and contamination controls per construct are specified in the evaluation protocol or supplementary evaluation specification. Table~\ref{tab:llm-pipeline} maps each construct through the full scoring pipeline.

\begin{table}[ht]
\centering
\caption{LLM evaluation pipeline: construct to dimension.
  Mode: L\,=\,likelihood-based (deterministic),
  G\,=\,generation-based (config-dependent), ECE = expected calibration error.
  Mechanisms A--D are detailed below under \emph{How outputs are scored}.}
\label{tab:llm-pipeline}
\vspace{0.4em}
\footnotesize
\resizebox{\columnwidth}{!}{%
\begin{tabular}{@{}llcll@{}}
\toprule
\textbf{Construct} & \textbf{Primary benchmark}
  & \textbf{Mode} & \textbf{Mechanism} & \textbf{Dimension}\\
\midrule
Factuality          & TruthfulQA \cite{lin2022truthfulqa}      & L / G & A.\ Reference; C.\ Judge & Safety\\
Reasoning (math)    & GSM8K \cite{cobbe2021gsm8k}              & G     & A.\ Reference            & Capability\\
Reasoning (general) & BIG-Bench Hard \cite{suzgun2023bbh}      & L     & B.\ Probability          & Capability\\
Knowledge           & MMLU \cite{hendrycks2021mmlu}            & L     & B.\ Probability          & Capability\\
Toxicity            & RealToxicityPrompts \cite{gehman2020rtp} & G     & C.\ Judge                & Safety\\
Adversarial         & HarmBench \cite{mazeika2024harmbench}    & G     & C.\ Judge                & Safety\\
Code                & HumanEval \cite{chen2021evaluating}      & G     & D.\ Execution            & Capability\\
Calibration         & ECE over MMLU \cite{guo2017calibration}  & L     & B.\ Probability          & Transparency\\
\bottomrule
\end{tabular}}
\end{table}

\paragraph{Availability of Log-Probabilities} Likelihood-based scoring requires access to token log probabilities, which depends on how the model is served. Open-weight models run locally expose the full logit vector. API-served models vary: some providers return top-$k$ log-probabilities per position, sufficient for likelihood ranking over answer options and for calibration error, but not for full-vocabulary perplexity; others expose no log-probabilities at all, making likelihood-based scoring unavailable. Where an instrument's designated mode is unavailable for a given model, the evaluation uses generation-based scoring as a fallback, the fallback is recorded in the evidence package, and likelihood-scored and generation-scored results obtained under the two modes are not pooled in comparative analyses. For models without log-probability access, calibration (ECE) is marked N/A unless a verbalized confidence protocol is used, in which case the protocol is recorded.

\paragraph{Output-Level Scoring Mechanisms}
\label{paragraph:scoring}
Every output-level instrument that scores the \emph{content or quality} of an output maps it to a quantitative score through one of four mechanisms. Three of them read the \emph{decoded text} and so are sensitive to decoding stochasticity; one reads the model's \emph{log-probabilities} and is deterministic given the model. Efficiency metrics are instead read from instrumentation, and the rule-based rate metrics (refusal, abstention) and the differential fairness metrics are computed over these outputs; all are catalogued in Table~\ref{tab:llm}. The shared primitive is the softmax: for logits $z$, the probability of token $i$ is
\begin{align}
p_i &= \frac{e^{z_i}}{\sum_j e^{z_j}}, \\
\log p(\text{token}) &= z_{\text{token}} - \log\!\sum_j e^{z_j}. \nonumber
\end{align}
Throughout, $\hat{y}$ is the candidate output and $y$ the reference.

\textbf{Reference Comparison.}
The output is compared to a reference by an overlap or similarity function. Exact match and token-$F_1$, with token sets $T(\cdot)$:
\begin{align}
\mathrm{EM} &= \frac{1}{N}\sum_{i=1}^{N}\mathbf{1}\!\left[\hat{y}_i = y_i\right],
\qquad F_1 = \frac{2PR}{P+R}, \\
P &=\frac{|T(\hat{y})\cap T(y)|}{|T(\hat{y})|},\qquad
R =\frac{|T(\hat{y})\cap T(y)|}{|T(y)|}. \nonumber
\end{align}
$n$-gram overlap (BLEU~\cite{papineni2002bleu}, with modified precision $p_n$, weights $w_n$, brevity penalty BP) and ROUGE-L~\cite{lin2004rouge} (from longest common subsequence):
\begin{align}
\mathrm{BLEU} &= \mathrm{BP}\cdot\exp\!\left(\sum_{n=1}^{N} w_n \log p_n\right), \\
\mathrm{BP} &=\begin{cases}1 & c>r\\ e^{1-r/c} & c\le r\end{cases} \nonumber \\
F_{\mathrm{LCS}} &=\frac{(1+\beta^2)PR}{R+\beta^2 P}. \nonumber
\end{align}
METEOR augments this $n$-gram matching with stemming and synonymy~\cite{banerjee2005meteor}. Embedding similarity (BERTScore~\cite{zhang2020bertscore}) matches each token to its closest counterpart by cosine similarity of contextual embeddings:
\begin{align}
R_{\mathrm{BERT}} &=\frac{1}{|y|}\sum_{y_i\in y}\max_{\hat{y}_j\in\hat{y}} y_i^{\top}\hat{y}_j, \\
F_{\mathrm{BERT}} &=\frac{2 P_{\mathrm{BERT}} R_{\mathrm{BERT}}}{P_{\mathrm{BERT}}+R_{\mathrm{BERT}}}. \nonumber
\end{align}

\textbf{Probability-Based Scoring.}
These operate directly on the model's token log-probabilities. Perplexity on a sequence of length $T$, and length-normalized likelihood ranking over candidate options $o$, are reference-free:
\begin{align}
\mathrm{PPL} &=\exp\!\left(-\frac{1}{T}\sum_{t=1}^{T}\log p(x_t\mid x_{<t})\right), \\
\mathrm{score}(o) &=\frac{1}{|o|}\sum_{t=1}^{|o|}\log p(o_t\mid \text{prompt},o_{<t}), \nonumber
\end{align}
with prediction $\hat{o}=\argmax_{o}\,\mathrm{score}(o)$. Expected calibration error additionally requires reference labels, since it compares confidence against realized accuracy~\cite{guo2017calibration}, over $M$ confidence bins $B_m$:
\begin{equation}
\mathrm{ECE}=\sum_{m=1}^{M}\frac{|B_m|}{n}\,\bigl|\mathrm{acc}(B_m)-\mathrm{conf}(B_m)\bigr|.
\end{equation}

\textbf{Model-Based Evaluation (LLM-as-Judge).}
A secondary model evaluates the candidate output. A trained classifier derives the relevant class probability from its softmax distribution; an NLI model averages entailment probability over $K$ claims against the source:
\begin{align}
\mathrm{tox} &=\softmax(z_\phi)_{\text{harmful}}, \\
\mathrm{Faith} &=\frac{1}{K}\sum_{k=1}^{K} p_{\mathrm{NLI}}(\text{entail}\mid \text{source},\text{claim}_k). \nonumber
\end{align}
An LLM judge~\cite{zheng2023mtbench} is the same mechanism applied to a general model. Pointwise, the score is the expectation over rating tokens $s\in\{1,\dots,S\}$, not the top token; pairwise, it reads the verdict-token probabilities, averaged over both presentation orders to remove position bias:
\begin{align}
\mathrm{score} &=\sum_{s=1}^{S} s\,p_{\mathrm{judge}}(s\mid \text{prompt},\hat{y}), \\
P(A\succ B) &=\tfrac{1}{2}\!\left[\frac{p(A\mid AB)}{p(A)+p(B)}+\frac{p(A\mid BA)}{p(A)+p(B)}\right]. \nonumber
\end{align}
Pairwise verdicts aggregate into a ranking by Bradley--Terry strengths $\beta$~\cite{bradley1952rank}, or by Elo update with step $K$~\cite{elo1978rating}:
\begin{align}
P(i\succ j) &=\sigma(\beta_i-\beta_j)=\frac{1}{1+e^{-(\beta_i-\beta_j)}}, \\
R_i &\leftarrow R_i + K\,(S_i-E_i), \nonumber \\
E_i &=\frac{1}{1+10^{(R_j-R_i)/400}}. \nonumber
\end{align}

\textbf{Execution-Based Evaluation.}
Behavior is verified rather than compared. With $n$ samples per problem and $c$ correct, the unbiased estimator of solving within $k$ attempts~\cite{chen2021evaluating}:
\begin{equation}
\mathrm{pass@}k = \mathbb{E}_{\text{problems}}\!\left[\,1-\frac{\binom{n-c}{k}}{\binom{n}{k}}\right].
\end{equation}

\paragraph{Aggregation and Uncertainty}
A per-item score becomes reportable only after aggregation with an uncertainty estimate. The mean over $N$ items is reported with a bootstrap confidence interval obtained by resampling the $N$ items $B$ times~\cite{efron1979bootstrap}:
\begin{equation}
\bar{x}=\frac{1}{N}\sum_{i=1}^{N} s_i \;\pm\; \mathrm{CI}_{\text{boot}}.
\end{equation}
Together with repeated runs across multiple seeds and appropriate significance testing, uncertainty estimation supports statistically defensible reporting. 

\subsubsection{Agentic and Multi-Agent Evaluation: Trajectory Level}
\label{sec:agentic}
\paragraph{Single-Agent Evaluation}
The unit of evaluation shifts from a single output to a complete \emph{trajectory} of actions. The trajectory-level instruments cover multi-step task completion, tool-use correctness, step and cost efficiency, error detection and recovery, goal fidelity, and oversight; full descriptors are in Table~\ref{tab:agent}. \emph{Cumulative} risk, where individually benign actions compound within a trajectory or across collaborating agents, is defined but scheduled for instrumentation in a later revision. The sandbox in which trajectories are executed and logged is defined in the toolkit.

The evaluation harness instrumented within the execution environment provides the components required to compute trajectory-level metrics: a \emph{programmatic task verifier} that checks whether the agent has achieved the goal state, using deterministic checkers for closed tasks and a fixed LLM judge for open ones~\cite{zhou2024webarena,liu2024agentbench}; an out-of-process \emph{action and trajectory logger} recording every tool call, argument, return value, and constraint-violation flag at each step~\cite{liu2024agentbench}; a \emph{fault injector} that introduces configurable failures mid-trajectory to test recovery behaviour~\cite{jia2026mas}; an \emph{interruption and override channel} delivering stop signals and human corrections independently of the agent's action loop~\cite{Zou2026WhenUC}; a \emph{per-step uncertainty probe} eliciting calibrated confidence estimates at each decision point~\cite{kadavath2022language,tomani2024uncertainty}; a \emph{pre-execution impact labeller} assigning each proposed action an impact level before dispatch, with high-impact or irreversible actions requiring an explicit confirmation event~\cite{jia2026mas}; and a \emph{structured evidence sink} assembling the full ordered sequence of observations, actions, and environment responses into the evidence package~\cite{liu2024agentbench}. The harness exposes an OpenAI-compatible interface and supports both API-served and local inference backends, so results are comparable across proprietary and open-weight models. Each trajectory is reproducible given five evaluation artifacts: environment version, task specification, agent configuration, decoding parameters, and random seed~\cite{zhou2024webarena}.

Scoring reuses the four output-level mechanisms: task success, tool-call accuracy, and pass@k are execution checks; open-ended goal verification and trajectory-optimality judgements are model-scored; the per-step uncertainty probe is probability arithmetic; and cost and consistency are read from instrumentation. The one metric specific to this evaluation level is step efficiency, the ratio of a reference-optimal path length to the agent's step count:
\begin{equation}
\eta_{\text{step}} = \min\!\left(1, \frac{L^{\ast}}{L}\right),\;
L=\text{steps},\; L^{\ast}=\text{optimal},
\end{equation}
capped at $1$ so detours are penalized and shortcuts cannot inflate the score. Because $\eta_{\text{step}}$ is bounded in $[0,1]$ and higher-is-better, it is banded under the higher-is-better rate family of Table~\ref{tab:metric-bands},
not the ratio-to-budget family.

\paragraph{Multi-Agent Evaluation}
When several agents collaborate on a task, failure modes arise that may not be observable at the single-agent level: miscoordination, redundant or conflicting actions, cascading errors across agents, and emergent unsafe behaviour. Recent MAS benchmarks therefore complement final task success with process-level evaluation. MultiAgentBench measures planning, communication, individual contribution, and milestone progress across different coordination topologies~\cite{zhu2025multiagentbench}, while Collab-Overcooked evaluates agents' ability to initiate and respond to collaboration~\cite{sun2025collab}. Failure analyses further identify inter-agent misalignment and incorrect verification or termination as failures that may remain hidden when only the final outcome is considered~\cite{cemri2026multi}. Process efficiency and scalability are also explicit concerns. For example, GEMMAS introduces Unnecessary Path Ratio (UPR) to quantify redundant reasoning paths~\cite{lee2025gemmas}, while SILO-BENCH introduces Relative Coordination Cost (RCC) to measure performance lost to coordination overhead relative to a single-agent baseline as team size increases~\cite{zhang2026silo}.

The framework therefore evaluates a multi-agent system at the level of the joint trajectory of all agents and their inter-agent messages. The MAS-specific instruments cover communication quality, delegation and role accuracy, coordination efficiency and reasoning redundancy, scalability with agent count, consensus and conflict resolution, error-propagation containment, verification and termination correctness, deadlock or livelock, and collective safety. Where applicable, evaluation also varies the agent roster or orchestration topology to test whether collaboration quality is stable across system configurations. 

Collective robustness is tested by introducing controlled faults to individual agents or communication channels. MAS-FIRE applies fault injection to both intra-agent reasoning and inter-agent coordination~\cite{jia2026mas}; the evaluation protocol records whether such faults are detected, contained, propagated, or recovered from. The trajectory-level harness is extended with the inter-agent message bus, role and subtask assignments, verification and termination events, and per-agent attribution of actions and constraint violations. Reproducibility additionally fixes the agent roster and topology, communication protocol, and orchestration policy. Scoring otherwise reuses the trajectory-level mechanisms above; inter-agent messages are scored for relevance and grounding by the judge mechanism of \emph{How outputs are scored}.

The multi-agent-specific instruments are listed at the foot of Table~\ref{tab:agent-mas}. One metric specific to this evaluation level is the cumulative unsafe-action rate. For a system of $M$ agents, where agent $i$ executes $T_i$ actions and $v_{i,t}\in\{0,1\}$ indicates whether action $t$ violates a specified safety constraint, \begin{equation} U_{\mathrm{MAS}} = \frac{\sum_{i=1}^{M}\sum_{t=1}^{T_i} v_{i,t}} {\sum_{i=1}^{M} T_i}. \end{equation} $U_{\mathrm{MAS}}$ is the fraction of actions in the joint trajectory that violate a safety constraint, irrespective of which agent originated them. It is a lower-is-better rate and a veto candidate, pending empirical validation. These instruments are defined here but not yet operationalized in version 1.0.

\subsubsection{Multimodal Evaluation: Cross-Modal Level}
\label{sec:multimodal}

The unit of evaluation is the relationship \emph{between} modalities. The cross-modal instruments cover grounding and cross-modal consistency, per-modality perception accuracy, modality-specific harms, and robustness to corrupted or adversarial inputs in each modality; full descriptors are in Table~\ref{tab:mm}.
Per-metric datasets are as follows: VQA accuracy is evaluated on VQAv2~\cite{goyal2017vqav2}, GQA~\cite{hudson2019gqa}, and MMBench~\cite{liu2024mmbench}; caption quality on MS-COCO~\cite{lin2014coco} and NoCaps~\cite{agrawal2019nocaps}; cross-modal retrieval on the MS-COCO retrieval split and Flickr30k~\cite{young2014flickr30k}; grounding accuracy on RefCOCO and RefCOCO+~\cite{yu2016refcoco}; OCR and text-read accuracy on TextVQA~\cite{singh2019textvqa}, DocVQA~\cite{mathew2021docvqa}, and ChartQA~\cite{masry2022chartqa}; ASR word error rate on LibriSpeech~\cite{panayotov2015librispeech} and FLEURS~\cite{conneau2023fleurs}; and cross-modal hallucination on POPE~\cite{li2023pope}, HallusionBench~\cite{guan2024hallusionbench}, and MMHal-Bench~\cite{sun2023mmhalbench}. 

Modality robustness is evaluated against a fixed perturbation set. Four graded corruption types are applied at three severity levels: Gaussian noise, blur, compression artifacts, and typographic attacks. Modality dropout (image-only and text-only ablations) is binary and is applied separately, since it admits no severity grading. Perturbed pairs are derived from the corresponding clean evaluation split.

Scoring reuses the output-level mechanisms: OCR, text-read accuracy, WER, and retrieval are reference comparisons; CLIPScore is embedding similarity; cross-modal hallucination is model-scored. Four metrics are specific to this evaluation level. VQA uses a consensus soft-accuracy over annotator agreement, where the denominator of three is the VQAv2 convention that an answer given by at least three of the ten independent annotators counts as fully correct~\cite{goyal2017vqav2}:
\begin{equation}
\mathrm{Acc}_{\mathrm{VQA}} = \frac{1}{N}\sum_{i=1}^{N}\min\!\left(1,\ \frac{\#\{\text{annotators giving }\hat{y}_i\}}{3}\right).
\end{equation}
Caption quality uses CIDEr, the weighted cosine similarity of TF--IDF $n$-gram vectors $\mathbf{g}^{n}$ against $M$ references, with SPICE as a scene-graph $F_1$ alternative~\cite{cider,spice}:
\begin{equation}
\mathrm{CIDEr} = \sum_{n=1}^{N} w_n\,\frac{1}{M}\sum_{j=1}^{M}\frac{\mathbf{g}^{n}(\hat{y})\cdot \mathbf{g}^{n}(r_j)}{\lVert\mathbf{g}^{n}(\hat{y})\rVert\,\lVert\mathbf{g}^{n}(r_j)\rVert}.
\end{equation}
Cross-modal retrieval uses Recall@$k$, the fraction of queries whose correct match falls in the top $k$:
\begin{equation}
\mathrm{R@}k = \frac{1}{N}\sum_{i=1}^{N}\mathbf{1}\!\left[\text{match}(q_i)\in \mathrm{top\text{-}}k(q_i)\right].
\end{equation}
Generation fidelity uses FID, the Fr\'echet distance between Gaussian fits to real and generated Inception features (the Inception Score is the alternative)~\cite{fid,inceptionscore}:
\begin{equation}
\mathrm{FID} = \lVert\mu_r-\mu_g\rVert_2^2 + \mathrm{Tr}\!\left(\Sigma_r+\Sigma_g-2(\Sigma_r\Sigma_g)^{1/2}\right).
\end{equation}

\subsection{Coverage Across System Types}
\label{sec:coverage}
\begin{table}[t]
\centering
\small
\setlength{\tabcolsep}{3pt}
\renewcommand{\arraystretch}{1}

\caption{Coverage of the eight spine dimensions across system types and the meta-evaluation layer.}
\label{tab:coverage}

\resizebox{\columnwidth}{!}{%
\begin{tabular}{L{1.9cm} L{3.1cm} L{3.6cm} L{3.3cm} L{2.5cm}}
\toprule
\textbf{Dimension} & \textbf{LLM (output)}
& \textbf{Agentic (trajectory)}
& \textbf{Multimodal (cross-modal)}
& \textbf{Meta-eval} \\
\midrule

Capability
& EM/F1, Pass@k, judge
& task success, tool-call, optimality, delegation$^{m}$, comm.\ quality$^{m}$
& VQA, caption, retrieval, grounding, OCR, WER
& N/A$^{a}$ \\

\cmidrule(lr){2-5}

Robustness
& self-consistency$^{\dagger}$
& error recovery, consistency, error-propagation containment$^{m}$, deadlock$^{m}$
& modality robustness
& test--retest stability \\

\cmidrule(lr){2-5}

Safety
& faithfulness, toxicity, refusal
& unsafe-action rate$^{\ddagger}$, cumulative unsafe-action (MAS)$^{m}$
& cross-modal hallucination$^{\ddagger}$
& N/A$^{a}$ \\

\cmidrule(lr){2-5}

Fairness
& bias/fairness gap
& fairness gap (outcomes)$^{s}$
& fairness gap (groups, accents)$^{s}$
& judge bias \\

\cmidrule(lr){2-5}

Transparency
& calibration (ECE)
& per-step calibration$^{s}$
& answer calibration$^{s}$
& judge--human agreement, inter-rater \\

\cmidrule(lr){2-5}

Governance
& contamination, format robustness, extraction validity,
model-governance completeness$^{s}$
& trajectory reproducibility, verifier validity
& per-modality contamination, annotation quality
& metric validity, discriminative power, contamination \\

\cmidrule(lr){2-5}

Oversight
& abstention, escalation
& interruptibility, override, confirmation-gating
& abstention, escalation$^{s}$; action-loop instruments N/A$^{b}$
& N/A$^{a}$ \\

\cmidrule(lr){2-5}

Efficiency
& latency, throughput, cost, memory, energy
& step efficiency, cost, coordination efficiency$^{m}$
& latency, cost, energy$^{s}$
& N/A$^{a}$ \\

\bottomrule
\end{tabular}%
}

$^{s}$ generic instrument applied across system types;
$^{\dagger}$ applicable but not yet operationalized at the output level;
$^{\ddagger}$ per-step or per-artifact only;
$^{a}$ not applicable to meta-evaluation;
$^{b}$ not applicable to systems without an action loop;
$^{m}$ multi-agent-system instrument (Table~\ref{tab:agent-mas}), not yet operationalized in version~1.0.

\end{table}

Generic instruments (fairness gap, calibration, efficiency) are specified once in
the LLM catalogue and applied across system types; the descriptor tables list
each once to avoid repetition. Cells marked $\dagger$, $\ddagger$, or $m$ are defined in the present framework but not yet operationalized in version 1.0; all are scheduled for the next revision. The dimensions marked N/A carry no instrument by construction and are not interpreted as failure.

\subsection{Score Normalization and Banding}
\label{sec:banding}
Every native metric score is mapped to a shared ordinal 0--4 band to support consistent interpretation across heterogeneous instruments without treating their native scales as directly equivalent. The mapping is governed by six families (Table~\ref{tab:metric-bands}), each with its own mapping rule. Every instrument in Appendix~\ref{app:catalog} is assigned to exactly one family;
where an instrument is not named in Table~\ref{tab:metric-bands}, the family
recorded in its descriptor row governs. Band thresholds are default framework thresholds;
thresholds may be revised between evaluation cycles through protocol versioning.

\begin{table}[t]
\centering
\scriptsize
\setlength{\tabcolsep}{2pt}
\renewcommand{\arraystretch}{0.92}

\caption{Metric families and band boundaries.}
\label{tab:metric-bands}

\begin{tabularx}{\columnwidth}{
@{}p{0.22\columnwidth}
   p{0.34\columnwidth}
   X@{}
}
\toprule
\textbf{Family} & \textbf{Metrics} & \textbf{Band rule} \\
\midrule

Higher-is-better
&
Accuracy, F1, Pass@$k$, VQA, Recall@$k$, task success,
tool accuracy, recovery, grounding, OCR, oversight metrics
&
0: $[0,.2)$;
1: $[.2,.4)$;
2: $[.4,.6)$;
3: $[.6,.8)$;
4: $[.8,1]$
\\
\midrule

Lower-is-better
&
Toxicity, unsafe-action, deadlock, hallucination,
contamination, WER
&
4: $[0,.05)$;
3: $[.05,.10)$;
2: $[.10,.20)$;
1: $[.20,.40)$;
0: $[.40,1]$
\\
\midrule

Lower-is-better continuous
&
ECE, FID, perplexity, fairness gap, robustness degradation,
judge bias, variance
&
$\mathrm{band}=4-\lfloor4v/B\rfloor$.
$B$: ECE=.20, FID=300, fairness=.20,
robustness=.30, bias=.15, variance=.10
\\
\midrule

Ratio-to-budget
&
Latency, throughput, cost, memory, energy,
coordination, token cost
&
For budget $B$:
4: $\le.25B$;
3: $\le.50B$;
2: $\le.75B$;
1: $\le B$;
0: $>B$
\\
\midrule

Target-band
&
Refusal, abstention, help-seeking
&
Inside $[lo,hi]$: 4; decreases with distance.
Defaults: unsafe [.85,.95], safe [0,.05],
abstention [.10,.30]
\\
\midrule

Comparative
&
LLM-judge win-rate, Elo,
trajectory optimality, Bradley--Terry
&
Pool quartiles:
top=4, Q3=3, Q2=2, bottom=1;
below floor=0
\\

\bottomrule
\end{tabularx}

\end{table}
\paragraph{Safety-Critical Overrides}
Three instruments carry hard ceilings that override normal banding.

\begin{description}
  \item[Contamination $\ge 0.40$]
  The affected benchmark is voided: its score is withdrawn, the construct it
  covered is marked N/A for that submission, and no dimension band may be
  computed from it.
 
  \item[Unsafe-action rate $\ge 0.20$]
  The Safety dimension is capped at band~0, and the system is flagged as unsuitable for deployment under the evaluated conditions.
 
  \item[Confirmation-gating $< 0.80$]
  The Oversight dimension is capped at band~1, reflecting insufficient confirmation for high-impact or irreversible actions.
\end{description}

\subsection{Meta-Evaluation Layer}
\label{sec:meta}
The meta-evaluation layer assesses the evaluation process rather than the model itself and provides evidence about the validity and reliability of reported results. Its full descriptors are in Table~\ref{tab:meta}. It covers four checks:

\begin{enumerate}[leftmargin=1.5em,itemsep=2pt,topsep=3pt]
  \item Construct validity of each instrument.
  \item Measurement reliability, inter-run and inter-rater.
  \item Data-contamination control.
  \item Statistical reporting: effect sizes and confidence intervals, rather than point estimates alone.
\end{enumerate}

\noindent Contamination and construct validity appear both here and under Governance (Section~\ref{sec:governance}). The division is by scope: the per-dataset and per-level checks are governance instruments owned by the
submitting evaluator, while the cross-cutting aggregates reported in Table~\ref{tab:meta} are computed once over the reference evaluation pool.

\subsection{Governance and Evidence Requirements}
\label{sec:governance}
Governance has two facets. Evaluation-governance instruments assess the evaluation process and differ by evaluation level, whereas model-governance instruments assess the governance artifacts accompanying the model and are shared across system types. At each level, the governance checks address distinct sources of score inflation and irreproducibility; the cross-cutting governance metrics (metric validity, discriminative power, sensitivity, and contamination) are aggregated in the meta-evaluation layer.

\begin{description}
  \item[LLM (output)] Benchmark contamination control per dataset~\cite{sainz2023,farooq2026evaluating}, plus score stability across prompt paraphrases, since a single template shifts both absolute scores and relative rankings~\cite{mizrahi2024}; answer extraction is validated against a hand-checked sample to ensure that parsing errors do not confound the measurement.
  \item[Agentic (trajectory)] Trajectory reproducibility from the five fixed evaluation artifacts (environment version, task specification, agent configuration, decoding parameters, and seed)~\cite{zhou2024webarena}, and task-verifier validity: the programmatic or LLM checker is itself audited against human judgements of goal completion~\cite{liu2024agentbench}. For multi-agent systems, per-agent attribution validity is added: the action and constraint-violation logger is audited to confirm that each event is assigned to the correct originating agent, so that cumulative and cross-agent metrics rest on a reliable attribution trail.
  \item[Multimodal (cross-modal)] Per-modality contamination control, with image, audio, and text overlap against training data checked separately, and reference-annotation quality, understood as the degree to which the characteristics of the data satisfy stated and implied needs when used under specified conditions~\cite{iso25024}, since automated caption and VQA metrics inherit the noise of their reference labels.
  \item[Meta] The cross-cutting governance metrics in Table~\ref{tab:meta}: construct validity, inter-run and inter-rater reliability, discriminative power, sensitivity, and contamination, each reported with effect sizes and confidence intervals rather than bare point scores.
\end{description}

\subsection{Independent Review and Process Safeguards}
\label{sec:safeguards}
Self-evaluation introduces a potential conflict of interest. Every reported profile is therefore accompanied by supporting evidence and checked through independent safeguards designed to reduce evaluator bias.

\begin{description}
  \item[Evidence submission] Each profile is accompanied by an evidence package: native scores, decoding configuration, seeds, bootstrap intervals, per-benchmark contamination checks, and the reference evaluation pool version against which any comparative metric was banded. The package schema is fixed in the toolkit. Reported results can therefore be independently inspected and verified.
  \item[Peer review] Before a submission enters the reference pool, it is reviewed by an independent evaluator, against the meta-evaluation checklist (Section~\ref{sec:meta}).
  \item[Independent re-runs] A 2\% sample of submissions is re-executed from the fixed evaluation artifacts by an independent evaluator. The sample is drawn to provide coverage across submitting organizations at least once per cycle, with the remainder weighted toward submissions carrying veto-adjacent scores. A re-run whose profile diverges beyond the reported confidence interval is flagged for resolution.
  \item[Dispute resolution] Disagreements among author, reviewer, and re-runner are escalated to the designated governance body and resolved on the basis of the recorded evidence, with the outcome logged.
\end{description}

\section{Alignment with the EU AI Act and the GPAI Code of Practice}
\label{sec:eu}

The proposed framework is designed to align with the EU regulatory framework and to support traceable and governance-relevant evaluation evidence. Providers of general-purpose AI (GPAI) models are subject to baseline obligations concerning technical documentation, information for downstream providers, copyright policies, and summaries of training-data content~\cite{aiact,ecGuidelines}. GPAI models classified as presenting systemic risk are subject to additional obligations, including model evaluation, adversarial testing, serious-incident reporting, and cybersecurity measures~\cite{aiact,codeofpractice}. The systemic-risk threshold is defined in terms of training compute above $10^{25}$ FLOPs or designation by the European Commission~\cite{ecGuidelines}.

\textbf{Applicability to evaluated systems.}
For systems that do not meet the systemic-risk criteria, the baseline GPAI obligations under Article~53 remain relevant, whereas the additional obligations under Article~55 apply to GPAI models with systemic risk. The framework nevertheless incorporates evaluation practices associated with Article~55, including model evaluation, documented adversarial testing, and risk assessment, in a proportionate manner based on system capability and deployment risk. This approach supports the generation of evaluation evidence that remains useful as system capabilities, deployment contexts, or regulatory requirements evolve.

\begin{figure}[t]
\centering
    \includegraphics[width=0.80\linewidth]{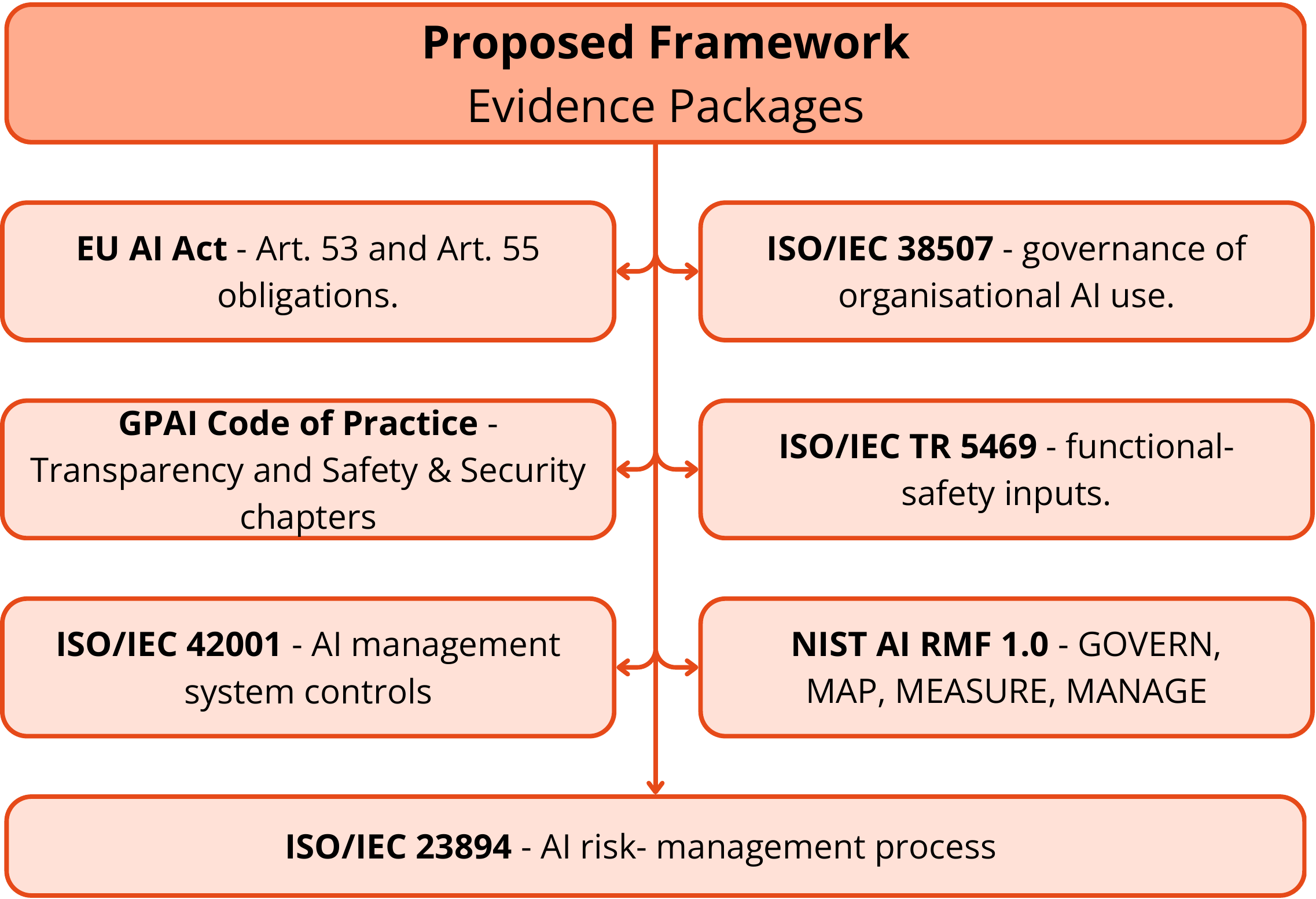}
\caption{Alignment of the proposed evaluation framework with the EU AI Act, the GPAI Code of Practice, and selected international AI governance, risk-management, and safety standards.}
\label{fig:alignment}
\end{figure}

The GPAI Code of Practice provides a structured means of supporting compliance with the relevant obligations under Articles~53 and~55. Its Safety and Security chapter emphasizes evaluations targeted to specific risk scenarios, appropriate scientific and technical rigor, and alignment with the model's expected use context~\cite{codeofpractice}. Figure~\ref{fig:alignment} summarizes the relationship between the proposed framework and relevant regulatory and standards-based frameworks, while Table~\ref{tab:eu-crosswalk} provides a provision-level crosswalk.

\begin{table}[t]
\centering
\scriptsize
\setlength{\tabcolsep}{2pt}
\renewcommand{\arraystretch}{0.95}

\caption{Crosswalk between framework elements, EU AI Act provisions, and the GPAI Code of Practice.}
\label{tab:eu-crosswalk}

\begin{tabularx}{\columnwidth}{
@{}p{0.22\columnwidth}
   p{0.38\columnwidth}
   X@{}
}
\toprule
\textbf{Element} & \textbf{EU AI Act} & \textbf{GPAI Code of Practice} \\
\midrule

Transparency
&
Art.~53(1)(a): technical documentation; Art.~53(1)(b): downstream information; Art.~53(1)(d): training-data summary; Annexes XI--XII
&
Transparency chapter; Measures~1.1--1.3 on model documentation, downstream information, quality/integrity, and retention
\\
\midrule

Safety
&
Art.~55(1)(a): model evaluation and adversarial testing; Annex XI, Sec.~2
&
Safety and Security chapter; model evaluation, red teaming, and Measure~10.1
\\
\midrule

Robustness
&
Art.~55(1)(a): state-of-the-art testing; Art.~55(1)(d): cybersecurity; Art.~51 and Annex XIII
&
Stress testing, resilience, robustness, and security measures
\\
\midrule

Governance / meta-evaluation
&
Art.~55(1)(a): standardized protocols and tools; Art.~56: Codes of Practice
&
State-of-the-art evaluation framework and rigor requirements
\\
\midrule

Oversight
&
Art.~55(1)(b): systemic-risk assessment and mitigation; Art.~55(1)(c): incident reporting; Art.~51 and Annex XIII
&
Ongoing monitoring, systemic-risk tracking, incident reporting, and corrective measures
\\
\midrule

Efficiency
&
Art.~53(1)(a), Annex XI, Sec.~1(2)(d)--(e): computational resources and energy use
&
Transparency chapter and Model Documentation Form provisions on compute and energy
\\
\midrule

Proportionality
&
Art.~51 and Annex XIII: systemic-risk classification and risk-based obligations
&
Proportionate requirements for systemic-risk, modified, and fine-tuned GPAI models
\\
\midrule

Evidence / audit trail
&
Art.~53(1)(a), Annexes XI--XII: documentation maintenance and availability
&
Up-to-date documentation and retention of relevant records
\\

\bottomrule
\end{tabularx}

\end{table}

\section{Alignment with ISO/IEC and NIST Frameworks}
\label{sec:standards}

Beyond the EU regulatory framework discussed in Section~\ref{sec:eu}, the proposed framework is designed so that its evaluation spine, evidence packages, and meta-evaluation layer can support established AI governance and risk-management standards. The following alignment is non-exhaustive.

\begin{description}[leftmargin=0pt,itemsep=5pt]

\item[ISO/IEC TS 5723 -- Trustworthiness vocabulary.]
ISO/IEC TS 5723:2022 defines trustworthiness as the ability to meet stakeholders' expectations in a verifiable manner and identifies characteristics whose relevance depends on the system, data, technology, processes, and application context~\cite{iso5723}. The proposed framework adopts this context-sensitive view of trustworthiness through the definitions introduced in Section~\ref{sec:introduction} and the evaluation dimensions in Table~\ref{tab:common_dimensions}.

\item[ISO/IEC 42001 -- AI management systems.]
ISO/IEC 42001:2023 specifies requirements for establishing, implementing, maintaining, and continually improving an AI management system (AIMS) for organisations that provide or use AI-based products or services. It addresses organisational context, leadership, planning, support, operation, performance evaluation, and continual improvement~\cite{iso42001}. The governance, meta-evaluation, and evidence-management components of the proposed framework can support performance evaluation and assurance activities within such an AI management system.

\item[ISO/IEC 23894 -- AI risk management.]
ISO/IEC 23894:2023 adapts established risk-management principles to AI and covers activities including communication, context establishment, risk assessment, risk treatment, monitoring, review, recording, and reporting~\cite{iso23894}. The framework's safety, robustness, fairness, transparency, oversight, and efficiency dimensions provide measurable evidence relevant to these risk-management activities, while the proportionality principle links evaluation depth to capability and deployment risk.

\item[ISO/IEC 38507 -- Governance of AI use.]
ISO/IEC 38507:2022 provides guidance for governing the organisational use of AI, including governance structures, policies, compliance, oversight, risk appetite, and risk-management controls~\cite{iso38507}. The governance and meta-evaluation components of the proposed framework support the operationalization of this guidance by producing structured evaluation profiles, documented protocols, and auditable evidence for oversight and decision-making.

\item[ISO/IEC TR 5469 -- Functional safety and AI.]
ISO/IEC TR~5469:2024 addresses the application of AI technologies in safety-related systems, including relevant properties, functional-safety risk factors, methods, constraints, and challenges~\cite{isotr5469}. The proposed framework is not itself a functional-safety standard; however, trajectory-level measures such as unsafe-action rate, confirmation-gating, and interruptibility can provide quantitative evidence relevant to functional-safety analysis where applicable.

\item[NIST AI RMF 1.0.]
The NIST AI Risk Management Framework 1.0 provides guidance for managing AI risks and promoting trustworthy AI throughout design, development, deployment, and use. Its trustworthiness characteristics include validity and reliability, safety, security and resilience, accountability and transparency, explainability and interpretability, privacy enhancement, and fairness with harmful bias managed. These characteristics are organized around four functions: GOVERN, MAP, MEASURE, and MANAGE~\cite{nistairmf}. The proposed framework supports these functions through its governance and meta-evaluation layer (GOVERN), coverage matrix and multidimensional profiles (MAP), metric catalogue and scoring procedures (MEASURE), and proportionality and decision policies (MANAGE).

\end{description}
\section{Limitations}
The framework has limits it states openly. Benchmark scores saturate and can be gamed, so a high number is evidence of capability under test conditions, not a guarantee of deployment behaviour. Coverage is incomplete for novel agentic and multimodal failure modes that current instruments were not designed to catch. The conflict of interest inherent in self-evaluation is mitigated by the process safeguards of  Section~\ref{sec:safeguards} but not eliminated, since independent re-runs cover only a sample. These limits are the reason the framework reports an uncertainty-bearing profile rather than a single pass mark, and why the meta-evaluation layer is part of the framework rather than an afterthought.



\section{Conclusion}
In this article, we presented a unified evaluation framework for LLMs, agentic and multi-agent systems, and multimodal models. The framework organizes evaluation across eight trustworthiness dimensions while preserving the specific requirements of output-level, trajectory-level, and cross-modal assessment. We described common scoring and reporting procedures and introduced a meta-evaluation layer to assess the quality of the evaluation itself. We also mapped the framework's evaluation and evidence requirements to relevant AI governance frameworks, standards, and regulations.

Evaluating AI trustworthiness remains challenging as system capabilities, deployment contexts, and failure modes evolve. The proposed framework provides a common basis for organizing evaluation evidence and supporting informed oversight. However, its practical value requires validation across diverse systems and use cases. Future work should focus on calibrating scoring thresholds based on use-cases, examining the relationship between evaluation profiles and deployment outcomes, and extending assessment methods to emerging agentic and multimodal risks.

\ifanon\else
\section*{Acknowledgment}
Resources used in preparing this research were provided, in part, by the Province of Ontario and the Government of Canada through CIFAR, as well as companies sponsoring the Vector Institute (\url{http://www.vectorinstitute.ai/\#partners}).
This research was funded by the European Union's Horizon Europe research and innovation programme under the AIXPERT project (Grant Agreement No.\ 101214389), which aims to develop an agentic, multi-layered, GenAI-powered framework for creating explainable, accountable, and transparent AI systems.
\fi

\bibliographystyle{IEEEtran}
\bibliography{references}

@misc{aiact,
  author = {{European Parliament and Council of the European Union}},
  title = {Regulation (EU) 2024/1689 of the European Parliament and of the Council of 13 June 2024 laying down harmonised rules on artificial intelligence (Artificial Intelligence Act)},
  year = {2024},
  howpublished = {OJ L, 2024/1689, 12 July 2024},
  note = {ELI: \url{http://data.europa.eu/eli/reg/2024/1689/oj}}
}

@article{raza2025humanibench,
  title={Humanibench: A human-centric framework for large multimodal models evaluation},
  author={Raza, Shaina and Narayanan, Aravind and Khazaie, Vahid Reza and Vayani, Ashmal and Radwan, Ahmed Y and Chettiar, Mukund S and Singh, Amandeep and Shah, Mubarak and Pandya, Deval},
  journal={ACM Transactions on Intelligent Systems and Technology},
  year={2025},
  publisher={ACM New York, NY}
}

@misc{eu_ai_act_2026,
  author       = {{European Parliament and Council of the European Union}},
  title        = {Regulation (EU) 2024/1689 Laying Down Harmonised Rules on Artificial Intelligence (Artificial Intelligence Act), Consolidated Version of 27 July 2026},
  year         = {2026},
  howpublished = {EUR-Lex},
  note         = {Consolidated text incorporating Regulation (EU) 2026/1744},
  url          = {https://eur-lex.europa.eu/eli/reg/2024/1689/2026-07-27/eng}
}

@inproceedings{sainz2023,
    title = "{NLP} Evaluation in trouble: On the Need to Measure {LLM} Data Contamination for each Benchmark",
    author = "Sainz, Oscar  and
      Campos, Jon  and
      Garc{\'i}a-Ferrero, Iker  and
      Etxaniz, Julen  and
      de Lacalle, Oier Lopez  and
      Agirre, Eneko",
    editor = "Bouamor, Houda  and
      Pino, Juan  and
      Bali, Kalika",
    booktitle = "Findings of the Association for Computational Linguistics: EMNLP 2023",
    month = dec,
    year = "2023",
    address = "Singapore",
    publisher = "Association for Computational Linguistics",
    url = "https://aclanthology.org/2023.findings-emnlp.722/",
    doi = "10.18653/v1/2023.findings-emnlp.722",
    pages = "10776--10787",
}

@article{mizrahi2024,
  title={State of what art? a call for multi-prompt llm evaluation},
  author={Mizrahi, Moran and Kaplan, Guy and Malkin, Dan and Dror, Rotem and Shahaf, Dafna and Stanovsky, Gabriel},
  journal={Transactions of the Association for Computational Linguistics},
  volume={12},
  pages={933--949},
  year={2024},
  publisher={MIT Press 255 Main Street, 9th Floor, Cambridge, Massachusetts 02142, USA~…}
}

@misc{codeofpractice,
  author       = {{European Commission}},
  title        = {The General-Purpose AI Code of Practice},
  year         = {2025},
  month        = jul,
  day          = {10},
  publisher    = {European Commission},
  url          = {https://digital-strategy.ec.europa.eu/en/policies/ai-code-practice},
  note         = {Published 10 July 2025}
}

@inproceedings{zhou2024webarena,
 author = {Zhou, Shuyan and Xu, Frank F and Zhu, Hao and Zhou, Xuhui and Lo, Robert and Sridhar, Abishek and Cheng, Xianyi and Ou, Tianyue and Bisk, Yonatan and Fried, Daniel and Alon, Uri and Neubig, Graham},
 booktitle = {International Conference on Learning Representations},
 editor = {B. Kim and Y. Yue and S. Chaudhuri and K. Fragkiadaki and M. Khan and Y. Sun},
 pages = {15585--15606},
 title = {WebArena: A Realistic Web Environment for Building Autonomous Agents},
 url = {https://proceedings.iclr.cc/paper_files/paper/2024/file/4410c0711e9154a7a2d26f9b3816d1ef-Paper-Conference.pdf},
 volume = {2024},
 year = {2024}
}

@inproceedings{liu2024agentbench,
 author = {Liu, Xiao and Yu, Hao and Zhang, Hanchen and Xu, Yifan and Lei, Xuanyu and Lai, Hanyu and Gu, Yu and Ding, Hangliang and Men, Kaiwen and Yang, Kejuan and Zhang, Shudan and Deng, Xiang and Zeng, Aohan and Du, Zhengxiao and Zhang, Chenhui and Shen, Sheng and Zhang, Tianjun and Su, Yu and Sun, Huan and Huang, Minlie and Dong, Yuxiao and Tang, Jie},
 booktitle = {International Conference on Learning Representations},
 editor = {B. Kim and Y. Yue and S. Chaudhuri and K. Fragkiadaki and M. Khan and Y. Sun},
 pages = {52989--53046},
 title = {AgentBench: Evaluating LLMs as Agents},
 url = {https://proceedings.iclr.cc/paper_files/paper/2024/file/e9df36b21ff4ee211a8b71ee8b7e9f57-Paper-Conference.pdf},
 volume = {2024},
 year = {2024}
}

@article{Zou2026WhenUC,
  title={When users change their mind: Evaluating interruptible agents in long-horizon web navigation},
  author={Zou, Henry Peng and Miao, Chunyu and Huang, Wei-Chieh and Chen, Yankai and Zhou, Yue and Zhang, Hanrong and Wu, Yaozu and Fang, Liancheng and Gu, Zhengyao and Zhang, Zhen and others},
  journal={arXiv preprint arXiv:2604.00892},
  year={2026}
}

@article{kadavath2022language,
  title={Language models (mostly) know what they know},
  author={Kadavath, Saurav and Conerly, Tom and Askell, Amanda and Henighan, Tom and Drain, Dawn and Perez, Ethan and Schiefer, Nicholas and Hatfield-Dodds, Zac and DasSarma, Nova and Tran-Johnson, Eli and others},
  journal={arXiv preprint arXiv:2207.05221},
  year={2022}
}

@article{jia2026mas,
  title={Mas-fire: Fault injection and reliability evaluation for llm-based multi-agent systems},
  author={Jia, Jin and Deng, Zhiling and Chen, Zhuangbin and Wang, Yingqi and Zheng, Zibin},
  journal={arXiv preprint arXiv:2602.19843},
  year={2026}
}

@article{tomani2024uncertainty,
  title={Uncertainty-based abstention in llms improves safety and reduces hallucinations},
  author={Tomani, Christian and Chaudhuri, Kamalika and Evtimov, Ivan and Cremers, Daniel and Ibrahim, Mark},
  journal={arXiv preprint arXiv:2404.10960},
  year={2024}
}

@inproceedings{papineni2002bleu, title = {{BLEU}: a Method for Automatic Evaluation of Machine Translation}, author = {Papineni, Kishore and Roukos, Salim and Ward, Todd and Zhu, Wei-Jing}, booktitle = {Proceedings of the 40th Annual Meeting of the Association for Computational Linguistics (ACL)}, pages = {311--318}, year = {2002}}

@inproceedings{lin2004rouge, title = {{ROUGE}: A Package for Automatic Evaluation of Summaries}, author = {Lin, Chin-Yew}, booktitle = {Text Summarization Branches Out}, pages = {74--81}, year = {2004}}

@inproceedings{banerjee2005meteor, title = {{METEOR}: An Automatic Metric for {MT} Evaluation with Improved Correlation with Human Judgments}, author = {Banerjee, Satanjeev and Lavie, Alon}, booktitle = {Proceedings of the ACL Workshop on Intrinsic and Extrinsic Evaluation Measures}, pages = {65--72}, year = {2005}}

@inproceedings{zhang2020bertscore, title = {{BERTScore}: Evaluating Text Generation with {BERT}}, author = {Zhang, Tianyi and Kishore, Varsha and Wu, Felix and Weinberger, Kilian Q. and Artzi, Yoav}, booktitle = {International Conference on Learning Representations (ICLR)}, year = {2020}}

@inproceedings{guo2017calibration, title = {On Calibration of Modern Neural Networks}, author = {Guo, Chuan and Pleiss, Geoff and Sun, Yu and Weinberger, Kilian Q.}, booktitle = {Proceedings of the 34th International Conference on Machine Learning (ICML)}, pages = {1321--1330}, year = {2017}}

@inproceedings{zheng2023mtbench, title = {Judging {LLM}-as-a-Judge with {MT}-Bench and Chatbot Arena}, author = {Zheng, Lianmin and Chiang, Wei-Lin and Sheng, Ying and others}, booktitle = {Advances in Neural Information Processing Systems (NeurIPS)}, year = {2023}}

@article{bradley1952rank, title = {Rank Analysis of Incomplete Block Designs: I. The Method of Paired Comparisons}, author = {Bradley, Ralph Allan and Terry, Milton E.}, journal = {Biometrika}, volume = {39}, number = {3/4}, pages = {324--345}, year = {1952}}

@book{elo1978rating, title = {The Rating of Chessplayers, Past and Present}, author = {Elo, Arpad E.}, publisher = {Arco}, year = {1978}}

@article{efron1979bootstrap, title = {Bootstrap Methods: Another Look at the Jackknife}, author = {Efron, Bradley}, journal = {The Annals of Statistics}, volume = {7}, number = {1}, pages = {1--26}, year = {1979}}

@inproceedings{lin2022truthfulqa, 
    title = "{T}ruthful{QA}: Measuring How Models Mimic Human Falsehoods",
    author = "Lin, Stephanie  and
      Hilton, Jacob  and
      Evans, Owain",
    editor = "Muresan, Smaranda  and
      Nakov, Preslav  and
      Villavicencio, Aline",
    booktitle = "Proceedings of the 60th Annual Meeting of the Association for Computational Linguistics (Volume 1: Long Papers)",
    month = may,
    year = "2022",
    address = "Dublin, Ireland",
    publisher = "Association for Computational Linguistics",
    url = "https://aclanthology.org/2022.acl-long.229/",
    doi = "10.18653/v1/2022.acl-long.229",
    pages = "3214--3252",
}

@inproceedings{li2023halueval,
    title = "{H}alu{E}val: A Large-Scale Hallucination Evaluation Benchmark for Large Language Models",
    author = "Li, Junyi  and
      Cheng, Xiaoxue  and
      Zhao, Xin  and
      Nie, Jian-Yun  and
      Wen, Ji-Rong",
    editor = "Bouamor, Houda  and
      Pino, Juan  and
      Bali, Kalika",
    booktitle = "Proceedings of the 2023 Conference on Empirical Methods in Natural Language Processing",
    month = dec,
    year = "2023",
    address = "Singapore",
    publisher = "Association for Computational Linguistics",
    url = "https://aclanthology.org/2023.emnlp-main.397/",
    doi = "10.18653/v1/2023.emnlp-main.397",
    pages = "6449--6464",
}

@article{cobbe2021gsm8k, 
  title={Training verifiers to solve math word problems},
  author={Cobbe, Karl and Kosaraju, Vineet and Bavarian, Mohammad and Chen, Mark and Jun, Heewoo and Kaiser, Lukasz and Plappert, Matthias and Tworek, Jerry and Hilton, Jacob and Nakano, Reiichiro and others},
  journal={arXiv preprint arXiv:2110.14168},
  year={2021}
}

@inproceedings{hendrycks2021math, 
  title={Measuring mathematical problem solving with the math dataset},
  author={Hendrycks, Dan and Burns, Collin and Kadavath, Saurav and Arora, Akul and Basart, Steven and Tang, Eric and Song, Dawn and Steinhardt, Jacob},
  journal={arXiv preprint arXiv:2103.03874},
  year={2021}
}

@inproceedings{suzgun2023bbh, 
    title = "Challenging {BIG}-Bench Tasks and Whether Chain-of-Thought Can Solve Them",
    author = {Suzgun, Mirac  and
      Scales, Nathan  and
      Sch{\"a}rli, Nathanael  and
      Gehrmann, Sebastian  and
      Tay, Yi  and
      Chung, Hyung Won  and
      Chowdhery, Aakanksha  and
      Le, Quoc  and
      Chi, Ed  and
      Zhou, Denny  and
      Wei, Jason},
    editor = "Rogers, Anna  and
      Boyd-Graber, Jordan  and
      Okazaki, Naoaki",
    booktitle = "Findings of the Association for Computational Linguistics: ACL 2023",
    month = jul,
    year = "2023",
    address = "Toronto, Canada",
    publisher = "Association for Computational Linguistics",
    url = "https://aclanthology.org/2023.findings-acl.824/",
    doi = "10.18653/v1/2023.findings-acl.824",
    pages = "13003--13051",
    }

@article{clark2018arc, 
  title={Think you have solved question answering? try arc, the ai2 reasoning challenge},
  author={Clark, Peter and Cowhey, Isaac and Etzioni, Oren and Khot, Tushar and Sabharwal, Ashish and Schoenick, Carissa and Tafjord, Oyvind},
  journal={arXiv preprint arXiv:1803.05457},
  year={2018}
}

@article{austin2021mbpp,
  title={Program synthesis with large language models},
  author={Austin, Jacob and Odena, Augustus and Nye, Maxwell and Bosma, Maarten and Michalewski, Henryk and Dohan, David and Jiang, Ellen and Cai, Carrie and Terry, Michael and Le, Quoc and others},
  journal={arXiv preprint arXiv:2108.07732},
  year={2021}
}

@inproceedings{hendrycks2021mmlu,
  title={Measuring massive multitask language understanding},
  author={Hendrycks, Dan and Burns, Collin and Basart, Steven and Zou, Andy and Mazeika, Mantas and Song, Dawn and Steinhardt, Jacob},
  journal={arXiv preprint arXiv:2009.03300},
  year={2020}
}

@inproceedings{gehman2020rtp, 
    title = "{R}eal{T}oxicity{P}rompts: Evaluating Neural Toxic Degeneration in Language Models",
    author = "Gehman, Samuel  and
      Gururangan, Suchin  and
      Sap, Maarten  and
      Choi, Yejin  and
      Smith, Noah A.",
    editor = "Cohn, Trevor  and
      He, Yulan  and
      Liu, Yang",
    booktitle = "Findings of the Association for Computational Linguistics: EMNLP 2020",
    month = nov,
    year = "2020",
    address = "Online",
    publisher = "Association for Computational Linguistics",
    url = "https://aclanthology.org/2020.findings-emnlp.301/",
    doi = "10.18653/v1/2020.findings-emnlp.301",
    pages = "3356--3369",}

@misc{iso42001,
  author       = {{ISO/IEC}},
  title        = {{ISO/IEC} 42001:2023 -- Information technology -- Artificial intelligence -- Management system},
  howpublished = {International Organization for Standardization, Geneva},
  year         = {2023},
  note         = {\url{https://www.iso.org/standard/42001}}
}

@misc{iso23894,
  author       = {{ISO/IEC}},
  title        = {{ISO/IEC} 23894:2023 -- Information technology -- Artificial intelligence -- Guidance on risk management},
  howpublished = {International Organization for Standardization, Geneva},
  year         = {2023},
  note         = {\url{https://www.iso.org/standard/77304.html}}
}

@misc{iso38507,
  author       = {{ISO/IEC}},
  title        = {{ISO/IEC} 38507:2022 -- Information technology -- Governance of {IT} -- Governance implications of the use of artificial intelligence by organizations},
  howpublished = {International Organization for Standardization, Geneva},
  year         = {2022},
  note         = {\url{https://www.iso.org/standard/56641.html}}
}

@misc{isotr5469,
  author       = {{ISO/IEC}},
  title        = {{ISO/IEC TR} 5469:2024 -- Artificial intelligence -- Functional safety and {AI} systems},
  howpublished = {International Organization for Standardization, Geneva},
  year         = {2024},
  note         = {\url{https://www.iso.org/standard/81283.html}}
}

@techreport{nistairmf,
  author      = {{National Institute of Standards and Technology}},
  title       = {Artificial Intelligence Risk Management Framework ({AI RMF} 1.0)},
  institution = {U.S. Department of Commerce},
  number      = {NIST AI 100-1},
  address     = {Gaithersburg, MD},
  year        = {2023},
  note        = {\url{https://nvlpubs.nist.gov/nistpubs/ai/nist.ai.100-1.pdf}}
}

@inproceedings{hartvigsen2022toxigen, 
    title = "{T}oxi{G}en: A Large-Scale Machine-Generated Dataset for Adversarial and Implicit Hate Speech Detection",
    author = "Hartvigsen, Thomas  and
      Gabriel, Saadia  and
      Palangi, Hamid  and
      Sap, Maarten  and
      Ray, Dipankar  and
      Kamar, Ece",
    editor = "Muresan, Smaranda  and
      Nakov, Preslav  and
      Villavicencio, Aline",
    booktitle = "Proceedings of the 60th Annual Meeting of the Association for Computational Linguistics (Volume 1: Long Papers)",
    month = may,
    year = "2022",
    address = "Dublin, Ireland",
    publisher = "Association for Computational Linguistics",
    url = "https://aclanthology.org/2022.acl-long.234/",
    doi = "10.18653/v1/2022.acl-long.234",
    pages = "3309--3326",}

@inproceedings{zhu2025multiagentbench,
  title={Multiagentbench: Evaluating the collaboration and competition of llm agents},
  author={Zhu, Kunlun and Du, Hongyi and Hong, Zhaochen and Yang, Xiaocheng and Guo, Shuyi and Wang, Daisy Zhe and Wang, Zhenhailong and Qian, Cheng and Tang, Xiangru and Ji, Heng and others},
  booktitle={Proceedings of the 63rd Annual Meeting of the Association for Computational Linguistics (Volume 1: Long Papers)},
  pages={8580--8622},
  year={2025}
}

@article{cemri2026multi,
  title={Why do multi-agent llm systems fail?},
  author={Cemri, Mert and Pan, Melissa Z and Yang, Shuyi and Agrawal, Lakshya A and Chopra, Bhavya and Tiwari, Rishabh and Keutzer, Kurt and Parameswaran, Aditya and Klein, Dan and Ramchandran, Kannan and others},
  journal={Advances in Neural Information Processing Systems},
  volume={38},
  year={2026}
}

@inproceedings{lee2025gemmas,
  title={Gemmas: Graph-based evaluation metrics for multi agent systems},
  author={Lee, Jisoo and Chang, Raeyoung and Kwon, Dongwook and Singh, Harmanpreet and Verma, Nikhil},
  booktitle={Proceedings of the 2025 Conference on Empirical Methods in Natural Language Processing: Industry Track},
  pages={1522--1532},
  year={2025}
}

@inproceedings{zhang2026silo,
  title={Silo-bench: A scalable environment for evaluating distributed coordination in multi-agent llm systems},
  author={Zhang, Yuzhe and Liu, Feiran and Shan, Yi and Huang, Xinyi and Yang, Xin and Zhu, Yueqi and Cheng, Xuxin and Liu, Cao and Zeng, Ke and Zhang, Terry Jingchen and others},
  booktitle={Proceedings of the 64th Annual Meeting of the Association for Computational Linguistics (Volume 1: Long Papers)},
  pages={29379--29398},
  year={2026}
}

@inproceedings{sun2025collab,
  title={Collab-overcooked: Benchmarking and evaluating large language models as collaborative agents},
  author={Sun, Haochen and Zhang, Shuwen and Niu, Lujie and Ren, Lei and Xu, Hao and Fu, Hao and Zhao, Fangkun and Yuan, Caixia and Wang, Xiaojie},
  booktitle={Proceedings of the 2025 Conference on Empirical Methods in Natural Language Processing},
  pages={4922--4951},
  year={2025}
}

@inproceedings{mazeika2024harmbench, 
  title={Harmbench: A standardized evaluation framework for automated red teaming and robust refusal},
  author={Mazeika, Mantas and Phan, Long and Yin, Xuwang and Zou, Andy and Wang, Zifan and Mu, Norman and Sakhaee, Elham and Li, Nathaniel and Basart, Steven and Li, Bo and others},
  journal={arXiv preprint arXiv:2402.04249},
  year={2024}
}

@misc{ecGuidelines,
  author       = {{European Commission}},
  title        = {Guidelines on the Scope of the Obligations for General-Purpose {AI} Models Established by Regulation (EU) 2024/1689 ({AI} Act)},
  howpublished = {European Commission, Brussels, C(2025) 5045 final},
  year         = {2025},
  note         = {Adopted 18 July 2025. \url{https://digital-strategy.ec.europa.eu/en/library/guidelines-scope-obligations-providers-general-purpose-ai-models-under-ai-act}}
}

@inproceedings{goyal2017vqav2, 
  title={Making the v in vqa matter: Elevating the role of image understanding in visual question answering},
  author={Goyal, Yash and Khot, Tejas and Summers-Stay, Douglas and Batra, Dhruv and Parikh, Devi},
  booktitle={Proceedings of the IEEE conference on computer vision and pattern recognition},
  pages={6904--6913},
  year={2017}
}

@inproceedings{hudson2019gqa, 
title={Gqa: A new dataset for real-world visual reasoning and compositional question answering},
  author={Hudson, Drew A and Manning, Christopher D},
  booktitle={2019 IEEE/CVF Conference on Computer Vision and Pattern Recognition (CVPR)},
  pages={6693--6702},
  year={2019},
  organization={IEEE}
}

@inproceedings{liu2024mmbench, 
  title={Mmbench: Is your multi-modal model an all-around player?},
  author={Liu, Yuan and Duan, Haodong and Zhang, Yuanhan and Li, Bo and Zhang, Songyang and Zhao, Wangbo and Yuan, Yike and Wang, Jiaqi and He, Conghui and Liu, Ziwei and others},
  booktitle={European conference on computer vision},
  pages={216--233},
  year={2024},
  organization={Springer}
}

@inproceedings{lin2014coco, 
  title={Microsoft coco: Common objects in context},
  author={Lin, Tsung-Yi and Maire, Michael and Belongie, Serge and Hays, James and Perona, Pietro and Ramanan, Deva and Doll{\'a}r, Piotr and Zitnick, C Lawrence},
  booktitle={European conference on computer vision},
  pages={740--755},
  year={2014},
  organization={Springer}
}

@inproceedings{agrawal2019nocaps,
  title={Nocaps: Novel object captioning at scale},
  author={Agrawal, Harsh and Desai, Karan and Wang, Yufei and Chen, Xinlei and Jain, Rishabh and Johnson, Mark and Batra, Dhruv and Parikh, Devi and Lee, Stefan and Anderson, Peter},
  booktitle={Proceedings of the IEEE/CVF international conference on computer vision},
  pages={8948--8957},
  year={2019}
}

@article{young2014flickr30k, 
  title={From image descriptions to visual denotations: New similarity metrics for semantic inference over event descriptions},
  author={Young, Peter and Lai, Alice and Hodosh, Micah and Hockenmaier, Julia},
  journal={Transactions of the association for computational linguistics},
  volume={2},
  pages={67--78},
  year={2014}
}

@inproceedings{yu2016refcoco, 
  title={Modeling context in referring expressions},
  author={Yu, Licheng and Poirson, Patrick and Yang, Shan and Berg, Alexander C and Berg, Tamara L},
  booktitle={European conference on computer vision},
  pages={69--85},
  year={2016},
  organization={Springer}
}

@inproceedings{singh2019textvqa, 
  title={Towards vqa models that can read},
  author={Singh, Amanpreet and Natarajan, Vivek and Shah, Meet and Jiang, Yu and Chen, Xinlei and Batra, Dhruv and Parikh, Devi and Rohrbach, Marcus},
  booktitle={Proceedings of the IEEE/CVF conference on computer vision and pattern recognition},
  pages={8317--8326},
  year={2019}
}

@inproceedings{mathew2021docvqa, 
  title={Docvqa: A dataset for vqa on document images},
  author={Mathew, Minesh and Karatzas, Dimosthenis and Jawahar, CV},
  booktitle={Proceedings of the IEEE/CVF winter conference on applications of computer vision},
  pages={2200--2209},
  year={2021}
}

@inproceedings{masry2022chartqa, 
    title = "{C}hart{QA}: A Benchmark for Question Answering about Charts with Visual and Logical Reasoning",
    author = "Masry, Ahmed  and
      Long, Do Xuan  and
      Tan, Jia Qing  and
      Joty, Shafiq  and
      Hoque, Enamul",
    editor = "Muresan, Smaranda  and
      Nakov, Preslav  and
      Villavicencio, Aline",
    booktitle = "Findings of the Association for Computational Linguistics: ACL 2022",
    month = may,
    year = "2022",
    address = "Dublin, Ireland",
    publisher = "Association for Computational Linguistics",
    url = "https://aclanthology.org/2022.findings-acl.177/",
    doi = "10.18653/v1/2022.findings-acl.177",
    pages = "2263--2279",
}

@inproceedings{panayotov2015librispeech, 
  title={Librispeech: an asr corpus based on public domain audio books},
  author={Panayotov, Vassil and Chen, Guoguo and Povey, Daniel and Khudanpur, Sanjeev},
  booktitle={2015 IEEE international conference on acoustics, speech and signal processing (ICASSP)},
  pages={5206--5210},
  year={2015},
  organization={IEEE}
}

@inproceedings{conneau2023fleurs,
  title={Fleurs: Few-shot learning evaluation of universal representations of speech},
  author={Conneau, Alexis and Ma, Min and Khanuja, Simran and Zhang, Yu and Axelrod, Vera and Dalmia, Siddharth and Riesa, Jason and Rivera, Clara and Bapna, Ankur},
  booktitle={2022 IEEE Spoken Language Technology Workshop (SLT)},
  pages={798--805},
  year={2023},
  organization={IEEE}
}

@inproceedings{li2023pope, 
    title = "Evaluating Object Hallucination in Large Vision-Language Models",
    author = "Li, Yifan  and
      Du, Yifan  and
      Zhou, Kun  and
      Wang, Jinpeng  and
      Zhao, Xin  and
      Wen, Ji-Rong",
    editor = "Bouamor, Houda  and
      Pino, Juan  and
      Bali, Kalika",
    booktitle = "Proceedings of the 2023 Conference on Empirical Methods in Natural Language Processing",
    month = dec,
    year = "2023",
    address = "Singapore",
    publisher = "Association for Computational Linguistics",
    url = "https://aclanthology.org/2023.emnlp-main.20/",
    doi = "10.18653/v1/2023.emnlp-main.20",
    pages = "292--305",
}

@inproceedings{guan2024hallusionbench,
  title={Hallusionbench: an advanced diagnostic suite for entangled language hallucination and visual illusion in large vision-language models},
  author={Guan, Tianrui and Liu, Fuxiao and Wu, Xiyang and Xian, Ruiqi and Li, Zongxia and Liu, Xiaoyu and Wang, Xijun and Chen, Lichang and Huang, Furong and Yacoob, Yaser and others},
  booktitle={Proceedings of the IEEE/CVF conference on computer vision and pattern recognition},
  pages={14375--14385},
  year={2024}
}

@misc{zou2023universal,
      title={Universal and Transferable Adversarial Attacks on Aligned Language Models}, 
      author={Andy Zou and Zifan Wang and J. Zico Kolter and Matt Fredrikson},
      year={2023},
      eprint={2307.15043},
      archivePrefix={arXiv},
      primaryClass={cs.CL}
}

@article{sun2023mmhalbench, 
 title = "Aligning Large Multimodal Models with Factually Augmented {RLHF}",
    author = "Sun, Zhiqing  and
      Shen, Sheng  and
      Cao, Shengcao  and
      Liu, Haotian  and
      Li, Chunyuan  and
      Shen, Yikang  and
      Gan, Chuang  and
      Gui, Liangyan  and
      Wang, Yu-Xiong  and
      Yang, Yiming  and
      Keutzer, Kurt  and
      Darrell, Trevor",
    editor = "Ku, Lun-Wei  and
      Martins, Andre  and
      Srikumar, Vivek",
    booktitle = "Findings of the Association for Computational Linguistics: ACL 2024",
    month = aug,
    year = "2024",
    address = "Bangkok, Thailand",
    publisher = "Association for Computational Linguistics",
    url = "https://aclanthology.org/2024.findings-acl.775/",
    doi = "10.18653/v1/2024.findings-acl.775",
    pages = "13088--13110",
    }

@inproceedings{cider, title = {{CIDEr}: Consensus-Based Image Description Evaluation}, author = {Vedantam, Ramakrishna and Zitnick, C. Lawrence and Parikh, Devi}, booktitle = {IEEE Conference on Computer Vision and Pattern Recognition (CVPR)}, pages = {4566--4575}, year = {2015}}

@inproceedings{spice, title = {{SPICE}: Semantic Propositional Image Caption Evaluation}, author = {Anderson, Peter and Fernando, Basura and Johnson, Mark and Gould, Stephen}, booktitle = {European Conference on Computer Vision (ECCV)}, pages = {382--398}, year = {2016}}

@inproceedings{fid, 
 author = {Heusel, Martin and Ramsauer, Hubert and Unterthiner, Thomas and Nessler, Bernhard and Hochreiter, Sepp},
 booktitle = {Advances in Neural Information Processing Systems},
 editor = {I. Guyon and U. Von Luxburg and S. Bengio and H. Wallach and R. Fergus and S. Vishwanathan and R. Garnett},
 pages = {},
 publisher = {Curran Associates, Inc.},
 title = {GANs Trained by a Two Time-Scale Update Rule Converge to a Local Nash Equilibrium},
 url = {https://proceedings.neurips.cc/paper_files/paper/2017/file/8a1d694707eb0fefe65871369074926d-Paper.pdf},
 volume = {30},
 year = {2017}}

@inproceedings{inceptionscore, 
 author = {Salimans, Tim and Goodfellow, Ian and Zaremba, Wojciech and Cheung, Vicki and Radford, Alec and Chen, Xi and Chen, Xi},
 booktitle = {Advances in Neural Information Processing Systems},
 editor = {D. Lee and M. Sugiyama and U. Luxburg and I. Guyon and R. Garnett},
 pages = {},
 publisher = {Curran Associates, Inc.},
 title = {Improved Techniques for Training GANs},
 url = {https://proceedings.neurips.cc/paper_files/paper/2016/file/8a3363abe792db2d8761d6403605aeb7-Paper.pdf},
 volume = {29},
 year = {2016}
 }

@inproceedings{wang2024mmlupro,
author = {Wang, Yubo and Ma, Xueguang and Zhang, Ge and Ni, Yuansheng and Chandra, Abhranil and Guo, Shiguang and Ren, Weiming and Arulraj, Aaran and He, Xuan and Jiang, Ziyan and Li, Tianle and Ku, Max and Wang, Kai and Zhuang, Alex and Fan, Rongqi and Yue, Xiang and Chen, Wenhu},
 booktitle = {Advances in Neural Information Processing Systems},
 doi = {10.52202/079017-3018},
 editor = {A. Globerson and L. Mackey and D. Belgrave and A. Fan and U. Paquet and J. Tomczak and C. Zhang},
 pages = {95266--95290},
 publisher = {Curran Associates, Inc.},
 title = {MMLU-Pro: A More Robust and Challenging Multi-Task Language Understanding Benchmark},
 url = {https://proceedings.neurips.cc/paper_files/paper/2024/file/ad236edc564f3e3156e1b2feafb99a24-Paper-Datasets_and_Benchmarks_Track.pdf},
 volume = {37},
 year = {2024}
}

@misc{chen2021evaluating,
  title={Evaluating large language models trained on code},
  author={Chen, Mark and Tworek, Jerry and Jun, Heewoo and Yuan, Qiming and Pinto, Henrique Ponde De Oliveira and Kaplan, Jared and Edwards, Harri and Burda, Yuri and Joseph, Nicholas and Brockman, Greg and others},
  journal={arXiv preprint arXiv:2107.03374},
  year={2021}
}

@article{farooq2026evaluating,
  title={Evaluating and regulating agentic ai: A study of benchmarks, metrics, and regulation},
  author={Farooq, Azib and Raza, Shaina and Karim, Nazmul and Iqbal, Hasan and Vasilakos, Athanasios V and Emmanouilidis, Christos},
  journal={Information Fusion},
  pages={104444},
  year={2026},
  publisher={Elsevier}
}

@techreport{iso5723,
  author      = {{ISO/IEC}},
  title       = {{ISO/IEC TS 5723:2022 -- Trustworthiness -- Vocabulary}},
  institution = {International Organization for Standardization},
  address     = {Geneva, Switzerland},
  type        = {Technical Specification},
  number      = {ISO/IEC TS 5723:2022},
  year        = {2022},
  month       = jul,
  url         = {https://www.iso.org/obp/ui/#iso:std:iso-iec:ts:5723:ed-1:v1:en}
}

@techreport{iso7498part2,
  author      = {{ISO}},
  title       = {{ISO 7498-2:1989 -- Information Processing Systems -- Open Systems Interconnection -- Basic Reference Model -- Part 2: Security Architecture}},
  institution = {International Organization for Standardization},
  address     = {Geneva, Switzerland},
  type        = {International Standard},
  number      = {ISO 7498-2:1989},
  year        = {1989},
  month       = feb,
  url         = {https://www.iso.org/obp/ui/es/#iso:std:iso:7498:-2:ed-1:v1:en}
}

@techreport{iso17572part1,
  author      = {{ISO}},
  title       = {{ISO 17572-1:2022 -- Intelligent Transport Systems (ITS) -- Location Referencing for Geographic Databases -- Part 1: General Requirements and Conceptual Model}},
  institution = {International Organization for Standardization},
  address     = {Geneva, Switzerland},
  type        = {International Standard},
  number      = {ISO 17572-1:2022},
  year        = {2022},
  month       = jul,
  edition     = {3},
  url         = {https://www.iso.org/obp/ui/#iso:std:iso:17572:-1:ed-3:v1:en}
}

@techreport{iso22989,
  author      = {{ISO/IEC}},
  title       = {{ISO/IEC 22989:2022 -- Information Technology -- Artificial Intelligence -- Artificial Intelligence Concepts and Terminology}},
  institution = {International Organization for Standardization},
  address     = {Geneva, Switzerland},
  type        = {International Standard},
  number      = {ISO/IEC 22989:2022},
  year        = {2022},
  month       = jul,
  edition     = {1},
  url         = {https://www.iso.org/obp/ui/#iso:std:iso-iec:22989:ed-1:v1:en}
}

@techreport{iso23643,
  author      = {{ISO/IEC}},
  title       = {{ISO/IEC 23643:2020 -- Software and Systems Engineering -- Capabilities of Software Safety and Security Verification Tools}},
  institution = {International Organization for Standardization},
  address     = {Geneva, Switzerland},
  type        = {International Standard},
  number      = {ISO/IEC 23643:2020},
  year        = {2020},
  month       = jun,
  edition     = {1},
  url         = {https://www.iso.org/obp/ui#iso:std:iso-iec:23643:dis:ed-1:v1:en}
}

@techreport{iso2382,
  author      = {{ISO/IEC}},
  title       = {{ISO/IEC 2382:2015 -- Information Technology -- Vocabulary}},
  institution = {International Organization for Standardization},
  address     = {Geneva, Switzerland},
  type        = {International Standard},
  number      = {ISO/IEC 2382:2015},
  year        = {2015},
  month       = may,
  edition     = {1},
  url         = {https://www.iso.org/obp/ui/#iso:std:iso-iec:2382:ed-1:v2:en}
}

@techreport{iso25024,
  author      = {{ISO/IEC}},
  title       = {{ISO/IEC 25024:2015 -- Systems and Software Engineering -- Systems and Software Quality Requirements and Evaluation (SQuaRE) -- Measurement of Data Quality}},
  institution = {International Organization for Standardization},
  address     = {Geneva, Switzerland},
  type        = {International Standard},
  number      = {ISO/IEC 25024:2015},
  year        = {2015},
  month       = oct,
  edition     = {1},
  url         = {https://www.iso.org/obp/ui/#iso:std:iso-iec:25024:ed-1:v1:en}
}

@techreport{iso27000,
  author      = {{ISO/IEC}},
  title       = {{ISO/IEC 27000:2018 -- Information Technology -- Security Techniques -- Information Security Management Systems -- Overview and Vocabulary}},
  institution = {International Organization for Standardization},
  address     = {Geneva, Switzerland},
  type        = {International Standard},
  number      = {ISO/IEC 27000:2018},
  year        = {2018},
  month       = feb,
  edition     = {5},
  url         = {https://www.iso.org/obp/ui/#iso:std:iso-iec:27000:ed-5:v1:en}
}

\appendices

\section{Metric Descriptor Catalog}
\label{app:catalog}

\noindent\textbf{Legend.}\enspace
\textit{Basis}: ref = reference-based, rf = reference-free, judge, behav = behavioral.\enspace
\textit{Scale}: bin, ord, cont, ratio.\enspace
\textit{Form}: point, rate, distrib, comp = comparative, corr = correlation.\enspace
\textit{Dir}: \hib\ higher-better, \lib\ lower-better, \tb\ target-band.\enspace
\textit{Norm} maps native output onto the shared 0--4 band using one of six families
(codes shown in the column):\enspace
\textsf{H}\,=\,higher-is-better rate
  (0:\,$[0,.20)$\; 1:\,$[.20,.40)$\; 2:\,$[.40,.60)$\;
   3:\,$[.60,.80)$\; 4:\,$[.80,1]$);\enspace
\textsf{L}\,=\,lower-is-better rate
  (4:\,$[0,.05)$\; 3:\,$[.05,.10)$\; 2:\,$[.10,.20)$\;
   1:\,$[.20,.40)$\; 0:\,$[.40,1]$);\enspace
\textsf{C($B$)}\,=\,lower-is-better continuous
  ($\mathrm{band}=4-\lfloor 4v/B\rfloor$);\enspace
\textsf{R}\,=\,ratio-to-budget
  (4:\,$v\!\le\!.25B$\;\dots\; 0:\,$v\!>\!B$; partner-declared $B$);\enspace
\textsf{W[$lo$,$hi$]}\,=\,target-band window
  (inside\,=\,4; linear decay outside);\enspace
\textsf{Q}\,=\,comparative quartile rank within the shared pool.
Veto overrides are marked \textbf{(v)}.

\begin{table*}[!p]
\scriptsize\centering\sloppy
\setlength{\tabcolsep}{2pt}
\renewcommand{\arraystretch}{1}\setlength{\emergencystretch}{2em}
\caption{LLM (output-level) metric descriptors}\label{tab:llm}
\begin{tabular}{L{2.0cm} L{1.45cm} L{1.1cm} L{1.0cm} L{0.9cm} c L{2.6cm} c L{2.3cm} L{2.5cm}}
\toprule
\textbf{Metric} & \textbf{Dim} & \textbf{Basis} & \textbf{Scale} & \textbf{Form} & \textbf{Dir} & \textbf{Native output} & \textbf{Norm} & \textbf{Data need} & \textbf{Failure signal}\\
\midrule
Accuracy / Exact Match & Capability & ref & bin\arr cont & rate & \hib & \% matching gold & H & reference labels & wrong final answers\\
F1 & Capability & ref & cont & point & \hib & 0--1 overlap & H & gold spans & partial / missed matches\\
Pass@k (code) & Capability & ref (exec) & bin\arr rate & rate & \hib & \% solved in $k$ & H & test suite / env & code fails tests\\
BLEU & Capability & ref & cont & point & \hib & 0--1 n-gram precision & H & reference text & low overlap w/ ref\\
ROUGE & Capability & ref & cont & point & \hib & 0--1 recall overlap & H & reference summary & missing ref content\\
METEOR & Capability & ref & cont & point & \hib & 0--1 stem / synonym-aware & H & reference text & semantic mismatch\\
BERTScore & Capability & ref (embed) & cont & point & \hib & 0--1 embed similarity & H & reference text & semantic divergence\\
LLM-judge / win-rate / Elo & Capability & judge & ord / comp & comp & \hib & win \% or Elo & Q & judge + opponent set & judged worse than base\\
Perplexity & Capability & rf & ratio & point & \lib & $\exp$(avg NLL) & C$^\ast$ & held-out text & poor LM fit\\
ECE (calibration) & Transparency & ref & cont & distrib & \lib & 0--1 conf--acc gap & C & labels + confidences & over- / under-confident\\
Faithfulness rate & Safety & judge / NLI & cont & rate & \hib & \% claims supported & H & source doc & unsupported claims\\
Toxicity & Safety & rf (clf) & cont & rate & \lib & 0--1 prob / \% toxic & L & none (classifier) & harmful output\\
Refusal rate & Safety & rule / judge & cont & rate & \tb & \% prompts refused & W & safe / unsafe prompt set & over- or under-refusal\\
Bias / fairness gap & Fairness & comp & cont & comp & \lib & delta across groups & C & group-labeled data & disparate performance\\
Abstention / deferral rate & Oversight & rule / judge & cont & rate & \tb & \% uncertain cases declined & W & answerable / unanswerable set & answers when it shouldn't\\
Escalation appropriateness & Oversight & judge / ref & cont & rate & \hib & precision--recall of ``needs human'' & H & gold escalation labels & misses review / false alarms\\
Inference latency & Efficiency & rf & ratio & point & \lib & sec / response & R & instrumentation & too slow for use case\\
Throughput & Efficiency & rf & ratio & rate & \hib & tokens or queries / sec & R & instrumentation & low serving capacity\\
Cost per query & Efficiency & rf & ratio & point & \lib & \$ per request & R & pricing + usage & over budget\\
Memory footprint & Efficiency & rf & ratio & point & \lib & params / VRAM & R & model spec & won't fit target hardware\\
Energy / carbon per query & Efficiency & rf & ratio & point & \lib & kWh or gCO$_2$e per request & R & power / carbon instrumentation & excess energy / emissions\\
\bottomrule
\multicolumn{10}{l}{\footnotesize $^\ast$ Perplexity comparable only within identical tokenizer / corpus.}\\
\end{tabular}
\end{table*}

\begin{table*}[!p]
\scriptsize\centering\sloppy
\setlength{\tabcolsep}{2pt}
\renewcommand{\arraystretch}{1.15}\setlength{\emergencystretch}{2em}
\caption{Agent (trajectory-level) metric descriptors: single-agent instruments)}\label{tab:agent}
\begin{tabular}{L{1.9cm} L{1.45cm} L{0.9cm} L{1.05cm} L{0.9cm} c L{2.15cm} L{2.6cm} L{1.95cm} L{2.1cm}}
\toprule
\textbf{Metric} & \textbf{Dim} & \textbf{Basis} & \textbf{Scale} & \textbf{Form} & \textbf{Dir} & \textbf{Native output} & \textbf{Norm (family [cuts])} & \textbf{Data need} & \textbf{Failure signal}\\
\midrule
Task success rate & Capability & behav & bin\arr rate & rate & \hib & \% goal achieved & H [0/.20/.40/.60/.80] & goal oracle / env & goal not reached\\
Sub-goal / progress & Capability & behav & cont & rate & \hib & \% sub-goals done & H [0/.20/.40/.60/.80] & annotated checkpoints & stalls partway\\
Pass@k (agent) & Capability & behav & bin\arr rate & rate & \hib & \% solved in $k$ tries & H [0/.20/.40/.60/.80] & env / goal oracle & fails across all $k$\\
Tool-call accuracy & Capability & behav & bin\arr rate & rate & \hib & \% correct tool+args & H [0/.20/.40/.60/.80] & gold trajectory / env & wrong tool / bad args\\
Step efficiency & Efficiency & behav & cont & comp & \hib & $\min(1, L^{\ast}/L)$ & H [0/.20/.40/.60/.80] & optimal path & redundant steps\\
Trajectory optimality & Capability & judge / behav & ord / cont & comp & \hib & path quality vs ref & Q (pool quartile) & ref path or judge & convoluted path\\
Error recovery rate & Robustness & behav & bin\arr rate & rate & \hib & \% failures recovered & H [0/.20/.40/.60/.80] & env w/ injected faults & can't recover\\
Cost (tokens / \$ / latency) & Efficiency & rf & ratio & point / rate & \lib & tokens, \$, seconds & R (partner budget) & instrumentation & over budget / slow\\
Unsafe action rate & Safety & rule / behav & bin\arr rate & rate & \lib & \% steps violating & L [.40/.20/.10/.05/0] \textbf{(v)} ${\ge}0.20$ & constraint spec / monitor & dangerous actions\\
Consistency & Robustness & rf & cont & distrib & \lib & variance across runs & C ($B{=}0.10$) & repeated runs & unstable outcomes\\
Interruptibility / stop-success & Oversight & behav & bin\arr rate & rate & \hib & \% clean halts on stop & H [0/.20/.40/.60/.80] & env w/ stop injection & ignores / mishandles stop\\
Human-override success & Oversight & behav & bin\arr rate & rate & \hib & \% overrides applied & H [0/.20/.40/.60/.80] & env w/ interventions & discards human input\\
Appropriate help-seeking & Oversight & judge / behav & cont & rate & \tb & \% asks only when needed & W [.10,.30] & tasks w/ need points & never asks / asks constantly\\
Deferral-on-uncertainty & Oversight & behav & bin\arr rate & rate & \hib & \% low-conf steps escalated & H [0/.20/.40/.60/.80] & confidence-instrumented env & barrels ahead when unsure\\
Confirmation-gating & Oversight & rule / behav & bin\arr rate & rate & \hib & \% high-impact actions gated & H [0/.20/.40/.60/.80] \textbf{(v)} ${<}0.80$ & action-impact spec & irreversible action w/o sign-off\\
\bottomrule
\end{tabular}
\end{table*}
\begin{table*}[!p]
\centering
\footnotesize
\setlength{\tabcolsep}{3pt}
\renewcommand{\arraystretch}{0.82}

\caption{Multi-agent trajectory-level metric descriptors.}
\label{tab:agent-mas}

\begin{tabularx}{\textwidth}{
@{}p{0.18\textwidth}
   p{0.15\textwidth}
   p{0.30\textwidth}
   X@{}
}
\toprule
\textbf{Metric / Dim.} &
\textbf{Type} &
\textbf{Output / Band} &
\textbf{Evidence / Failure} \\
\midrule

Communication quality \textit{(Capability)}
&
judge; cont.; rate; $\uparrow$
&
\% relevant/grounded msgs.; H [.20/.40/.60/.80]
&
Message log + rubric; off-topic/ungrounded communication
\\

Delegation accuracy \textit{(Capability)}
&
behav.; bin.; rate; $\uparrow$
&
\% subtasks assigned correctly; H [.20/.40/.60/.80]
&
Role assignment; incorrect agent selected
\\

Coordination efficiency \textit{(Efficiency)}
&
behav.; ratio; comp.; $\downarrow$
&
Msgs./steps relative to optimum; R (budget)
&
Reference trace; redundant/duplicated work
\\

Conflict resolution \textit{(Robustness)}
&
behav.; bin.; rate; $\uparrow$
&
\% conflicts resolved; H [.20/.40/.60/.80]
&
Induced disagreement; incorrect reconciliation
\\

Error containment \textit{(Robustness)}
&
behav.; bin.; rate; $\uparrow$
&
\% injected faults contained; H [.20/.40/.60/.80]
&
Fault injection; cascading failure
\\

Deadlock/livelock \textit{(Robustness)}
&
behav.; bin.; rate; $\downarrow$
&
\% stalled/looping runs; L [.40/.20/.10/.05/0]
&
Long-horizon tasks; indefinite stall/loop
\\

Verification correctness \textit{(Robustness)}
&
behav.; bin.; rate; $\uparrow$
&
\% outcomes verified; H [.20/.40/.60/.80]
&
Verification oracle; incorrect verification
\\

Termination correctness \textit{(Robustness)}
&
behav.; bin.; rate; $\uparrow$
&
\% correctly terminated; H [.20/.40/.60/.80]
&
Terminal oracle; premature/failed termination
\\

UPR \textit{(Efficiency)}
&
behav.; ratio; rate; $\downarrow$
&
Redundant reasoning fraction; L [.40/.20/.10/.05/0]
&
Communication graph; non-contributing paths
\\

RCC \textit{(Efficiency)}
&
comp.; ratio; rate; $\downarrow$
&
$\max(0,1-S_{\rm MAS}/S_1)$; L [.40/.20/.10/.05/0]
&
Single-agent baseline; coordination loss
\\

Unsafe-action rate \textit{(Safety)}
&
rule/behav.; bin.; rate; $\downarrow$
&
\% violating steps; L [.40/.20/.10/.05/0]
&
Constraint monitor; compounding unsafe actions
\\

\bottomrule
\end{tabularx}

\vspace{2pt}
{\scriptsize
\textit{Note:} Type = basis; scale; form; direction.
H/L/R = higher-is-better, lower-is-better, and ratio-to-budget families.
Metrics are defined but not yet operationalized in v1.0.}

\vspace{-4pt}
\end{table*}

\begin{table*}[!p]
\centering
\footnotesize
\setlength{\tabcolsep}{3pt}
\renewcommand{\arraystretch}{0.82}

\caption{Multimodal cross-modal metric descriptors.}
\label{tab:mm}

\begin{tabularx}{\textwidth}{
@{}p{0.18\textwidth}
   p{0.15\textwidth}
   p{0.30\textwidth}
   X@{}
}
\toprule
\textbf{Metric / Dim.} &
\textbf{Type} &
\textbf{Output / Band} &
\textbf{Evidence / Failure} \\
\midrule

VQA accuracy \textit{(Capability)}
&
ref.; bin.; rate; $\uparrow$
&
\% correct; H [.20/.40/.60/.80]
&
Q--A pairs; incorrect visual reasoning
\\

CIDEr/SPICE \textit{(Capability)}
&
ref.; cont.; point; $\uparrow$
&
Caption quality; H [.20/.40/.60/.80]
&
Reference captions; inaccurate captions
\\

Recall@$k$ \textit{(Capability)}
&
ref.; bin.; rate; $\uparrow$
&
\% correct in top-$k$; H [.20/.40/.60/.80]
&
Paired corpus; incorrect modality match
\\

CLIPScore \textit{(Capability)}
&
embed.; cont.; point; $\uparrow$
&
Image--text similarity; H [.20/.40/.60/.80]
&
Encoder; weak cross-modal alignment
\\

FID/IS \textit{(Capability)}
&
ref.; cont.; distrib.; $\downarrow$
&
Distance to real distribution; C ($B=300$)
&
Reference images; unrealistic generation
\\

Grounding accuracy \textit{(Capability)}
&
ref.; bin.; rate; $\uparrow$
&
\% correct regions; H [.20/.40/.60/.80]
&
Region labels; incorrect grounding
\\

OCR accuracy \textit{(Capability)}
&
ref.; cont.; rate; $\uparrow$
&
$1-$CER / \% correct; H [.20/.40/.60/.80]
&
Transcription; misread visual text
\\

WER \textit{(Capability)}
&
ref.; ratio; rate; $\downarrow$
&
Word error rate; L [.40/.20/.10/.05/0]
&
Transcript; ASR errors
\\

Cross-modal hallucination \textit{(Safety)}
&
judge; cont.; rate; $\downarrow$
&
\% unsupported claims; L [.40/.20/.10/.05/0]
&
Source modality; absent content described
\\

Modality robustness \textit{(Robustness)}
&
comp.; cont.; comp.; $\downarrow$
&
Perturbation degradation; C ($B=.30$)
&
Perturbed pairs; sensitivity to noise/dropout
\\

\bottomrule
\end{tabularx}

\vspace{2pt}
{\scriptsize
\textit{Note:} Type = basis; scale; form; direction.
H/L/C = higher-is-better, lower-is-better, and continuous families.}

\vspace{-4pt}
\end{table*}

\begin{table*}[!t]
\centering
\footnotesize
\setlength{\tabcolsep}{3pt}
\renewcommand{\arraystretch}{0.82}

\caption{Meta-evaluation metric descriptors.}
\label{tab:meta}

\begin{tabularx}{\textwidth}{
@{}p{0.18\textwidth}
   p{0.15\textwidth}
   p{0.30\textwidth}
   X@{}
}
\toprule
\textbf{Metric / Dim.} &
\textbf{Type} &
\textbf{Output / Band} &
\textbf{Evidence / Failure} \\
\midrule

Judge--human agreement \textit{(Transparency)}
&
comp.; cont.; corr.; $\uparrow$
&
Correlation with human ratings; H [.20/.40/.60/.80]
&
Human ratings; judge--human disagreement
\\

Inter-rater reliability \textit{(Transparency)}
&
ref.; cont.; distrib.; $\uparrow$
&
$\kappa$ / $\alpha$; H [.20/.40/.60/.80]
&
$\geq2$ raters; rater disagreement
\\

Metric validity \textit{(Governance)}
&
comp.; cont.; corr.; $\uparrow$
&
Correlation with criterion; H [.20/.40/.60/.80]
&
Criterion measure; metric tracks noise
\\

Discriminative power \textit{(Governance)}
&
ref.; cont.; distrib.; $\uparrow$
&
Score separation; H [.20/.40/.60/.80]
&
Multi-system scores; poor differentiation
\\

Test--retest stability \textit{(Robustness)}
&
ref.; cont.; distrib.; $\uparrow$
&
Rerun variance; C ($B=.10$)
&
Repeated runs; unstable scores
\\

Sensitivity \textit{(Governance)}
&
comp.; cont.; comp.; $\uparrow$
&
Detection of known differences; H [.20/.40/.60/.80]
&
Known-gap systems; missed differences
\\

Contamination/leakage \textit{(Governance)}
&
rule; bin.; rate; $\downarrow$
&
\% test items in training; L [.40/.20/.10/.05/0]
&
Train/test overlap; inflated scores
\\

Judge bias \textit{(Fairness)}
&
comp.; cont.; comp.; $\downarrow$
&
Score shift from artifact; C ($B=.15$)
&
Controlled probes; length/position/self bias
\\

\bottomrule
\end{tabularx}

\vspace{2pt}
{\scriptsize
\textit{Note:} Type = basis; scale; form; direction.
corr. = correlation; $\kappa$/$\alpha$ = agreement coefficients.}

\vspace{-4pt}
\end{table*}


\end{document}